\documentclass[runningheads]{llncs}

\usepackage{eccv}

\usepackage{eccvabbrv}

\usepackage{graphicx}
\usepackage{booktabs}
\usepackage{tabularx}

\usepackage{rotating}
\usepackage{makecell}
\usepackage[hidelinks]{hyperref}

\usepackage[accsupp]{axessibility}  

\usepackage{hyperref}

\usepackage{orcidlink}

\begin{document}

\title{RiO-DETR: DETR for Real-time Oriented Object Detection} 

\author{Zhangchi Hu\inst{1}\orcidlink{0009-0008-4030-825X} \and
Yifan Zhao\inst{1}\orcidlink{0009-0005-1301-3585} \and
Yansong Peng\inst{1}\orcidlink{0000-0003-3504-0611} \and
Wenzhang Sun\inst{3}\orcidlink{0000-0001-8924-8747} \and
Jie Chen\inst{1}\orcidlink{0009-0009-1809-2592} \and
Peixi Wu\inst{1}\orcidlink{0009-0003-1376-9628} \and
Xinghao Wang\inst{2} \and
Xiangchen Yin\inst{1}\orcidlink{0009-0006-7593-7473} \and
Dongsheng Jiang\inst{2}\orcidlink{0000-0002-7390-9173} \and
Hebei Li\inst{1}\dag\orcidlink{0000-0002-7529-6331} \and
Xiaoyan Sun\inst{1,4}\orcidlink{0000-0003-3638-5566}
}

\authorrunning{Z. Hu et al.}

\institute{University of Science and Technology of China \and
Huawei Technologies Co., Ltd.  \and
Tsinghua University \and
Institute of Artificial Intelligence, Hefei Comprehensive National
Science Center \\
\email{\{huzhangchi, lihebei\}@mail.ustc.edu.cn}, \email{sunxiaoyan@ustc.edu.cn} \\ 
\emph{$^\dag$Corresponding author}}

\maketitle

\vspace{-1.5em}

\begin{figure*}[h]
  \centering
  \includegraphics[height=4cm, width=\textwidth, keepaspectratio]{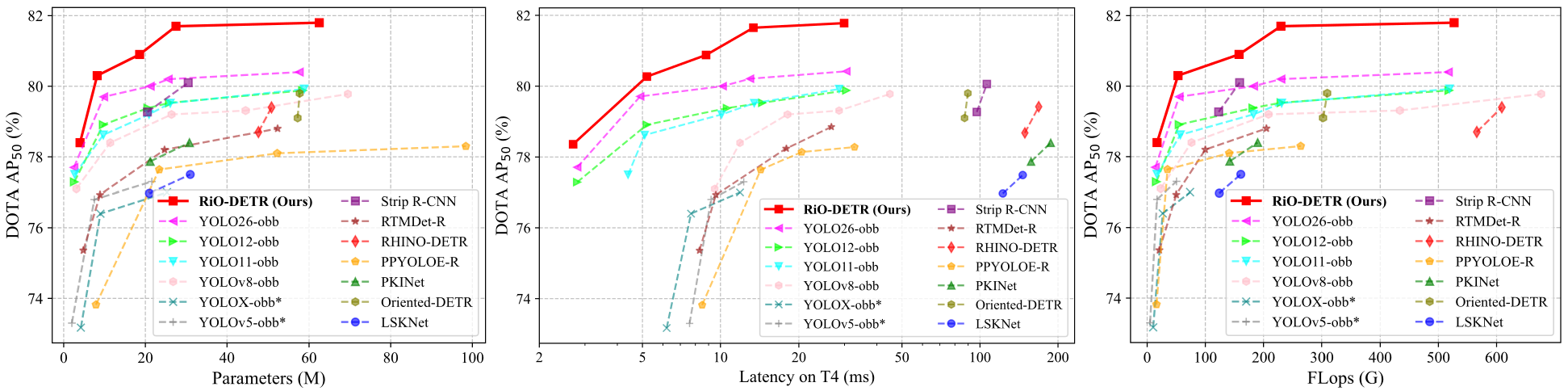}
  \caption{Comparisons with other detectors in terms of model size (left), latency (mid), and computational cost (right) on DOTA-1.0 under single-scale training and testing protocol. * denotes a community implemented version.}
  \label{fig:teaser}
\end{figure*}

\vspace{-1.5em}

\begin{abstract}

We present \textbf{RiO-DETR}: \textbf{DETR} for \textbf{R}eal-t\textbf{i}me \textbf{O}riented Object Detection, the first real-time oriented detection transformer to the best of our knowledge. Adapting DETR to oriented bounding boxes (OBBs) poses three challenges: semantics-dependent orientation, angle periodicity that breaks standard Euclidean refinement, and an enlarged search space that slows convergence. RiO-DETR resolves these issues with task-native designs while preserving real-time efficiency. First, we propose Content-Driven Angle Estimation by decoupling angle from positional queries, together with Rotation-Rectified Orthogonal Attention to capture complementary cues for reliable orientation. Second, Decoupled Periodic Refinement combines bounded coarse-to-fine updates with a Shortest-Path Periodic Loss for stable learning across angular seams. Third, Oriented Dense O2O injects angular diversity into dense supervision to speed up angle convergence at no extra cost. Extensive experiments on DOTA-1.0, DIOR-R, and FAIR-1M-2.0 demonstrate RiO-DETR establishes a new speed--accuracy trade-off for real-time oriented detection. GitHub Repository: \href{https://github.com/RicePasteM/RiO-DETR}{https://github.com/RicePasteM/RiO-DETR}.

  
\end{abstract}

\section{Introduction}
\label{sec:intro}


Oriented object detection extends horizontal bounding boxes (HBBs) to localize objects with arbitrary rotations, which is crucial for aerial imagery, remote sensing, and scene text understanding \cite{xia2018dota, ding2021object, cheng2022anchor, zhou2022mmrotate, girshick2015fast, ren2016faster, redmon2016you, tian2019fcos, zou2023object, sun2023neural, sun2026muse}. With the rise of edge computing, the bottleneck has evolved from mere detection accuracy to the speed-accuracy trade-off. This raises a critical question: how can we maintain high-quality oriented bounding box (OBBs) detection while meeting rigorous real-time performance requirements?


In the CNN paradigm, real-time oriented detectors have established robust baselines, especially oriented variants of YOLO and RTMDet-style frameworks \cite{jocher2022ultralytics, ge2021yolox, xu2022pp, lyu2022rtmdet, xie2021oriented, han2021align, wen2023comprehensive}. Meanwhile, DETR-style detectors have recently achieved real-time efficiency in horizontal detection by leveraging lightweight designs and efficient training strategies \cite{zhao2024detrs, chen2024lw, robinson2025rf, hu2025dome, peng2024d, huang2025deim, liu2025hi}. However, real-time oriented DETRs remain under-explored. Existing oriented DETR variants are often burdened by heavy attention designs and complex sampling modules in pursuit of peak accuracy \cite{hu2023emo2, zhao2024orientedformer, zhao2024projecting, lee2025hausdorff, zeng2024ars, dai2022ao2}, which inherently limits their ability to match the high throughput of CNN-based counterparts.


This gap is not primarily due to insufficient model capacity; rather, it reflects inherent architectural bottlenecks that emerge when adapting DETR from HBB to OBB. We identify three key bottlenecks that systematically limit real-time oriented DETRs: \textbf{(1) Semantic--Geometric Coupling and Feature Collapse.} Unlike the standard $(cx, cy, w, h)$ format, the orientation $\theta$ is strongly driven by semantic appearance cues such as texture flow and dominant axes. Encoding $\theta$ solely as a geometric prior in positional queries can introduce noise and misguide the attention mechanism. Furthermore, naively aligning attention with the major axis risks feature collapse, where lateral structures are insufficiently attended. \textbf{(2) Periodicity Mismatch in Angle Refinement.} Standard DETR decoders typically refine bounding boxes through Euclidean additive updates, such as the inverse-sigmoid formulation. Applying such updates to the cyclic angular domain introduces discontinuities at periodic boundaries, leading to unstable gradients and unreliable refinement \cite{chen2020piou, yang2021learning, yang2021rethinking, yang2022kfiou, yang2020arbitrary}. \textbf{(3) Slow Convergence in the Expanded Search Space.} OBBs introduce additional degrees of freedom, which significantly expands the search space for bipartite matching and slows convergence. While techniques like dense supervision or one-to-many training are effective for HBB detection, they often lack sufficient angular diversity to accelerate orientation learning in the OBB setting.


These observations motivate a fundamental shift in design: achieving real-time oriented detection requires moving beyond simply appending an angle branch or increasing computational overhead. Instead, we focus on reformulating core components to natively handle oriented geometry. To this end, we present \textbf{RiO-DETR}, a \textbf{R}eal-t\textbf{i}me \textbf{O}riented detection transformer that effectively bridges the gap between accuracy and latency. Built upon a delicately implemented baseline, RiO-DETR introduces three task-specific designs to resolve the aforementioned bottlenecks:
\begin{itemize}
    \item To decouple semantic and geometric cues, we introduce \textbf{Content-Driven Angle Estimation}. This approach employs a \textbf{Geometry-Decoupled Query Encoding} that facilitates angle prediction by leveraging semantic context rather than relying on rigid geometric priors. To address feature collapse with minimal overhead, we further propose \textbf{Rotation-Rectified Orthogonal Attention}, which captures both axial and lateral cues to enhance orientation inference.
    \item To address periodicity issues, we introduce \textbf{Decoupled Periodic Refinement}. Specifically, we replace standard Euclidean updates with a bounded coarse-to-fine periodic mechanism and a \textbf{Shortest-path Periodic $L_1$ Loss}. This approach enables stable optimization across angular boundaries and resolves refinement artifacts caused by periodic discontinuities.
    \item To address slow convergence, we develop \textbf{Oriented Dense O2O}, a training strategy that injects angular diversity by applying independent random rotations to stitched image quadrants, significantly accelerating the convergence of angle predictions.
\end{itemize}


\textbf{RiO-DETR achieves a superior speed--accuracy balance} for real-time oriented detection (Fig.~\ref{fig:teaser}). On DOTA-1.0, RiO-DETR-n attains \textbf{78.4} AP$_{50}$ with only \textbf{2.7\,ms} end-to-end latency (TensorRT FP16 on NVIDIA T4). Meanwhile, RiO-DETR-x reaches \textbf{81.8} AP$_{50}$ at \textbf{29.9\,ms}. Both models outperform state-of-the-art real-time detectors at similar speeds. Our approach also shows consistent improvements on DIOR-R and FAIR-1M-2.0 across various scales. These results demonstrate that end-to-end transformers can be highly efficient for oriented object detection. We hope our design provides a robust framework and inspires further exploration in this field.

\section{Related Works}

\subsubsection{CNN-based Oriented Object Detection.} Early oriented detectors mainly extended horizontal CNN detectors with angle regression. Two-stage methods \cite{xie2021oriented, lin2017focal, yang2020rotated} (e.g., RoI Transformer \cite{ding2019learning}) achieve high precision via rotation-invariant RoI operations, while single-stage/anchor-free counterparts (e.g., S2ANet \cite{yujie2024s2anet}, FCOS-O \cite{tian2019fcos}) simplify the pipeline for higher efficiency. Recently, oriented YOLO variants \cite{jocher2022ultralytics, sohan2024review, khanam2024yolov11, tian2025yolov12, sapkota2025yolo26, ge2021yolox, xu2022pp} and RTMDet-R \cite{lyu2022rtmdet} have advanced real-time oriented detection. However, most still depend on dense prediction and heuristic anchors, typically requiring NMS, which complicates deployment and adds hyperparameter sensitivity. Moreover, lightweight designs for oriented detection \cite{cai2024poly, yuan2025strip, li2025lsknet} often introduce complex structures that raise compute/memory access costs, leading to noticeable inference latency.

\subsubsection{Real-time Object Detection.} While the YOLO \cite{jocher2022ultralytics} ecosystem long dominated real-time detection, the emergence of DETR \cite{carion2020end} introduced a NMS-free, end-to-end paradigm. Early DETR variants \cite{zhu2020deformable, meng2021conditional, liu2022dab, li2022dn, zhang2022dino} struggled with computational overhead, but recent models like RT-DETR \cite{zhao2024detrs, chen2024lw} have bridged this gap using efficient hybrid encoders and uncertainty-minimal query selection. Further advancements \cite{robinson2025rf, hu2025dome, lv2024rt, wang2025rt, liao2025rt}, such as D-FINE \cite{peng2024d} and DEIM \cite{huang2025deim, huang2025real}, have enhanced precision and accelerated convergence through refined optimization and dense supervision. These breakthroughs prove the feasibility of real-time DETRs, yet their success remains largely confined to horizontal bounding boxes.

\subsubsection{DETR-based Oriented Object Detection.} Motivated by the success of DETRs \cite{carion2020end}, several works have attempted to adapt the Transformer architecture for oriented bounding boxes \cite{zhao2024orientedformer}. Methods like AO2-DETR \cite{dai2022ao2} and EMO2-DETR \cite{hu2023emo2} introduce oriented proposal generation and rotation-aware attention mechanisms to align features with arbitrary object orientations. Recent state-of-the-art models, such as ARS-DETR \cite{zeng2024ars}, Oriented-DETR \cite{zhao2024projecting} and RHINO \cite{lee2025hausdorff}, further address the matching and representation problem of OBBs. Despite their strong performance, these methods predominantly build upon heavy architectures and prioritize precision over inference speed, excluding them from real-time applications. To the best of our knowledge, designing a DETR-based oriented detector that maintains high accuracy while meeting strict real-time constraints remains an unresolved challenge.

\section{Methods}

\subsection{Best of Both Worlds: Building a Strong Real-time Baseline} 

While numerous studies have migrated oriented object detection to the DETR family, most of them rely on DINO \cite{zhang2022dino} or Deformable-DETR \cite{zhu2020deformable}, which is not fully optimized for inference efficiency. RT-DETRv2 is originally designed for horizontal object detection and does not come with an oriented detection (OBB) implementation. Therefore, we first develop an oriented RT-DETRv2 baseline by introducing an OBB regression head and adapting the angle representation, losses, and denoising scheme following RHINO-DETR \cite{lee2025hausdorff}. To make the baseline strong and fair under real-time constraints, we further consolidate two complementary training designs that have proven effective for DETR-style detectors: the universal matching strategy from D-FINE and Dense O2O from DEIM \cite{huang2025deim}. Additional implementation details are provided in Appendix A.

Table \ref{tab:baseline_comparison} shows the comparison between our implemented baseline and previous works. Our implementation exhibits competitive performance, which forms a solid foundation for our method. However, there remains a noticeable gap compared with state-of-the-art counterparts (1.18 AP$_{50}$), indicating that simply porting existing DETR techniques is insufficient for closing the accuracy–efficiency gap in oriented detection. It is our task-oriented improvements that solve the rooted issues in DETR-based oriented object detection, and elevate RiO-DETR to state-of-the-art accuracy–efficiency performance while keeping the parameter count, FLOPs, and inference latency virtually unchanged.

\subsection{Content-Driven Angle Estimation}
\label{sec:orientation}

\begin{figure}[t]
  \centering
  \includegraphics[width=\linewidth]{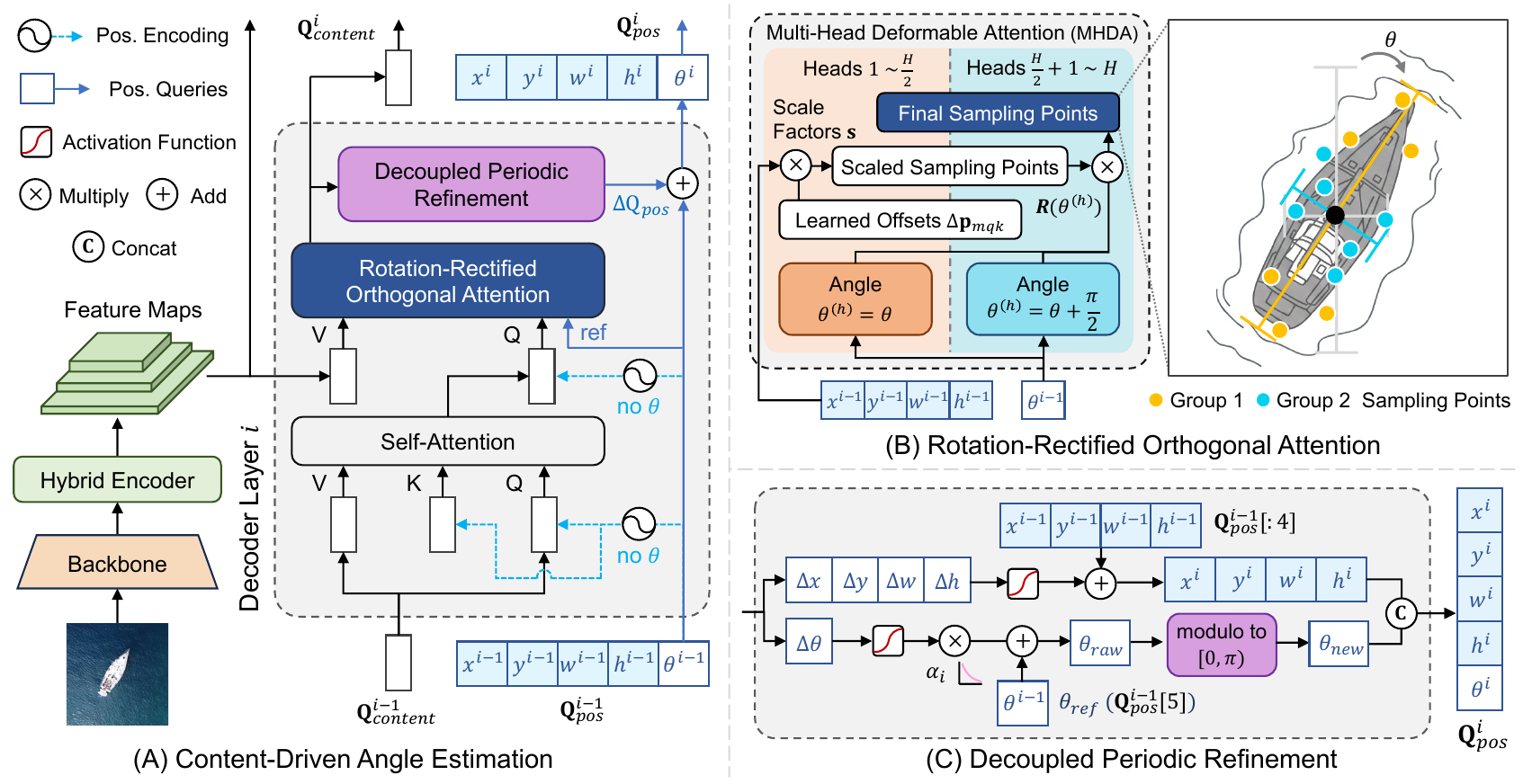}
  \caption{The main architecture of our proposed RiO-DETR. The framework highlights three key components: (A) Content-Driven Angle Estimation, (B) Rotation-Rectified Orthogonal Attention, and (C) Decoupled Periodic Refinement.}
  \label{fig:main}
\end{figure}

Existing DETR-based oriented object detectors treat the orientation $\theta$ as a geometric component symmetric to box coordinates $(c_x,c_y,w,h)$. Consequently, these methods \cite{hu2023emo2, zhao2024orientedformer, zhao2024projecting, lee2025hausdorff, zeng2024ars, dai2022ao2} intuitively embed the full 5-dimensional tuple $(cx, cy, w, h, \theta)$ into the query embeddings. This joint design implicitly assumes that spatial localization and angular rotation can be optimized in a homogeneous Euclidean manner, letting the decoder refine the entire geometric state as a unified representation. 

However, unlike $(c_x,c_y)$, the OBB orientation $\theta$ is not uniquely determined by geometric correspondence. 
It is periodic and admits equivalent parameterizations; for instance, $(w,h,\theta)$ and $(h,w,(\theta+\pi/2)\bmod \pi)$ describe the same physical rectangle.
As a result, $\theta$ is effectively a canonical choice induced by annotation conventions (e.g., long-side direction or heading) rather than a purely geometric quantity, and is often disambiguated only by appearance cues such as dominant axes, part layouts, texture flows, and semantic definitions.
Injecting $\theta$ into positional embeddings therefore imposes a rigid geometric prior that can be noisy during early training and potentially non-smooth around periodic boundaries, which may misguide attention sampling and interfere with content-driven refinement in the decoder.
We provide a more formal discussion from the viewpoint of quotient-consistency in Appendix~B.

Motivated by this asymmetry, we propose \textbf{Content-Driven Angle Estimation} (illustrated in Fig.~\ref{fig:main}(A)(B)).
Specifically, we decouple orientation from the positional prior via \textbf{Geometry-Decoupled Query Encoding}, encouraging the model to regress angles primarily from semantic context learned through content queries. Then, we design \textbf{Rotation-Rectified Orthogonal Attention} to extract rotation-aligned object features for more robust angle estimation.

\subsubsection{Geometry-Decoupled Query Encoding}
We formulate the object query $Q$ as a composition of a content part $Q_{content}$ and a positional part $Q_{pos}$. Unlike previous works, we strictly limit the positional embedding to the 4-dimensional spatial domain, explicitly excluding angular information.

Let $\mathbf{p}_{ref} \in \mathbb{R}^{N \times 5}$ denote the reference points containing $(cx, cy, w, h, \theta)$, where $N$ is the number of object queries. The positional embedding is generated via a coordinate encoder $\phi(\cdot)$, which consists of a sinusoidal positional encoding (PE) and a multi-layer perceptron (MLP):

\begin{equation}
    Q_{pos} = \phi(\mathbf{p}_{ref}[..., :4]) = \text{MLP}(\text{PE}(\mathbf{p}_{ref}[..., :4]))
\end{equation}

where $\mathbf{p}_{ref}[..., :4]$ represents only the spatial coordinates $(cx, cy, w, h)$. By masking the $\theta$ dimension during embedding generation, we ensure the positional queries remain rotation-invariant.

Simultaneously, the orientation information is latently modeled within the learnable content embeddings $Q_{content}$. This forces the decoder to extract rotation cues from the image features such as texture direction and object heading.

\subsubsection{Rotation-Rectified Orthogonal Attention}

While decoupling angle from positional embedding aids convergence, precise feature extraction requires sampling locations that align with the object's geometry. Previous oriented DETRs uniformly align all attention heads with the object's major axis, which leads to a feature collapse where the model disproportionately focuses on longitudinal details while neglecting lateral structural information.

To address this, we impose an orthogonal constraint on the multi-head attention mechanism to ensure feature coverage along both the major and minor axes of the oriented bounding box. For a query with predicted orientation $\theta$, we divide the total attention heads $H$ (assuming $H$ is an even integer) into two distinct groups. The sampling rotation angle $\theta^{(h)}$ for the $h$-th head, where $h \in \{1, 2, \dots, H\}$, is defined as:

\begin{equation}
    \theta^{(h)} = 
    \begin{cases} 
    \theta & \text{if } h \le \frac{H}{2} \\
    \theta + \frac{\pi}{2} & \text{if } h > \frac{H}{2}
    \end{cases}
\end{equation}

This formulation forces the first half of the heads to sample features aligned with the object's predicted heading, while the second half samples orthogonally. Notably, this design introduces no additional parameters or GFlops, achieving performance gains without extra computational overhead.

Let $\Delta \mathbf{p}_{qk} \in \mathbb{R}^2$ be the learned offset for the $k$-th sampling point of the $q$-th query, and $\mathbf{s}_q \in \mathbb{R}^2$ be its corresponding scale factor (width and height). The spatial center of the query is denoted as $\mathbf{p}_{q} = \mathbf{p}_{ref}[q, :2]$. The final sampling location $S$ is rectified by the head-specific rotation matrix $\mathbf{R}(\theta^{(h)})$:

\begin{equation}
    S(\mathbf{p}_q, \Delta \mathbf{p}_{qk}) = \mathbf{p}_{q} + \mathbf{R}(\theta^{(h)}) (\Delta \mathbf{p}_{qk} \odot \mathbf{s}_q)
\end{equation}

where $\odot$ denotes the element-wise multiplication, and $\mathbf{R}(\theta^{(h)})$ is the rotation matrix derived from the head-specific angle:

\begin{equation}
    \mathbf{R}(\theta^{(h)}) = 
    \begin{bmatrix} 
    \cos(\theta^{(h)}) & -\sin(\theta^{(h)}) \\
    \sin(\theta^{(h)}) & \cos(\theta^{(h)}) 
    \end{bmatrix}
\end{equation}

This orthogonal splitting strategy allows the model to explicitly disentangle features along the object's length and width, improving robustness for aspect-ratio and angle combined prediction.

\begin{figure}[t]
  \centering
  \includegraphics[width=\linewidth, keepaspectratio]{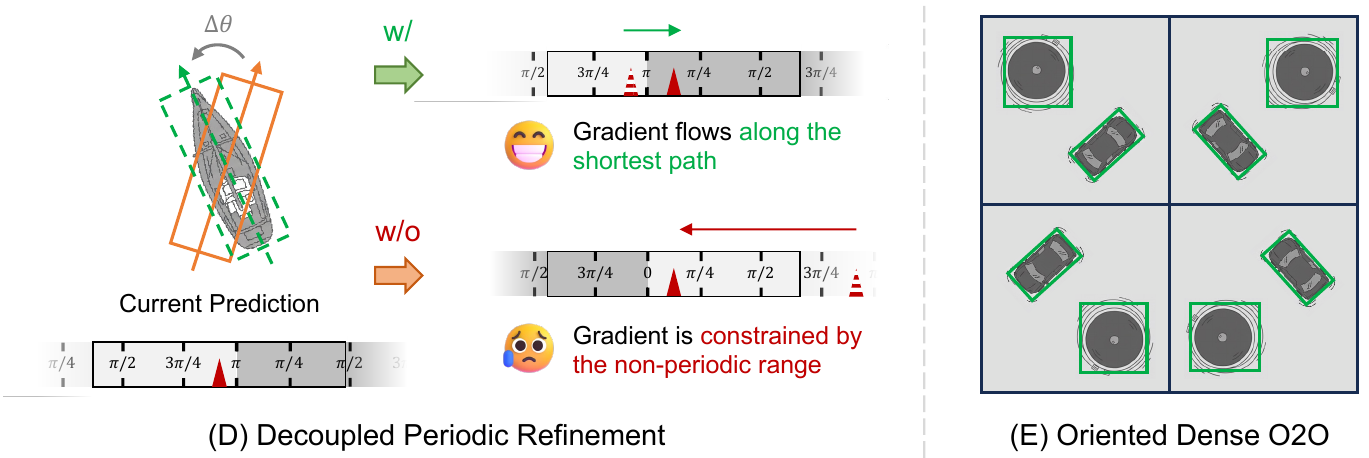}
  \caption{An intuitive illustration of (D) Decoupled Periodic Refinement and (E) Oriented Dense O2O.}
  \label{fig:main2}
\end{figure}

\subsection{Decoupled Periodic Refinement}

Standard DETR-style decoders refine boxes by iteratively adding predicted offsets in the inverse-sigmoid space. This update rule implicitly assumes that each regressed variable lies in a Euclidean and globally continuous domain, which holds well for spatial coordinates $(c_x,c_y,w,h)$. However, the orientation $\theta$ in OBB resides on a periodic space. Treating $\theta$ as an unconstrained Euclidean scalar, i.e., directly applying additive updates and standard $L_1$ regression induces a mismatched geometric assumption in decoder refinement: numerically distant angles can be geometrically adjacent across the boundary (e.g., $0 \leftrightarrow \pi$), leading to discontinuous gradients and unstable refinement near periodic seams.

To resolve this geometry mismatch, we introduce \textbf{Decoupled Periodic Refinement}, which jointly redesigns both the refinement update and the optimization metric for $\theta$. In other words, we make the decoder refinement consistent with the periodic topology of orientation, rather than merely adding an auxiliary trick. The overall workflow is illustrated in Fig. \ref{fig:main} (C) and \ref{fig:main2} (D).

For spatial dimensions, we retain the standard inverse-sigmoid update. Given the predicted spatial offset $\Delta \mathbf{b}$ and the reference box spatial parameters $\mathbf{b}_{ref}$, the refined spatial coordinates are computed as
$\sigma(\Delta \mathbf{b} + \sigma^{-1}(\mathbf{b}_{ref}))$, where $\sigma(\cdot)$ denotes the sigmoid function. For orientation, we adopt a bounded coarse-to-fine periodic update that explicitly respects the cyclic domain. Let $\Delta \theta_i$ be the raw angle offset predicted by the $i$-th decoder layer ($i\in\{1,\dots,L\}$) and $\theta_{ref}$ be the reference angle. We first bound the update magnitude with $\tanh(\cdot)$ and scale it using a layer-wise decaying factor $\alpha_i$:
\begin{equation}
\alpha_i = \alpha_0^{-i},
\end{equation}
where $\alpha_0>1$ controls the decay rate. This design enforces a coarse-to-fine refinement schedule: early layers perform larger corrective rotations, while later layers are restricted to fine-grained tuning, improving stability under periodicity. The bounded update is then applied as
\begin{equation}
\theta_{raw} = \theta_{ref} + \tanh(\Delta \theta_i)\cdot \alpha_i.
\end{equation}
Finally, we map the result back onto the canonical angular domain $[0,\pi)$ via periodic normalization:
\begin{equation}
\theta_{new} =
\begin{cases}
(\theta_{raw} \bmod \pi) + \pi & \text{if } (\theta_{raw} \bmod \pi) < 0 \\
(\theta_{raw} \bmod \pi) & \text{otherwise.}
\end{cases}
\end{equation}

Crucially, a periodic update alone is insufficient if the optimization objective still measures distance in a Euclidean sense. A standard $L_1$ loss over-penalizes boundary-adjacent cases, producing gradients that point along the longer arc and thus contradict the periodic refinement behavior. Therefore, we additionally replace the angular regression metric with a \textbf{Shortest-path Periodic $L_1$ Loss}:
\begin{equation}
\mathcal{L}_{angle} = \min\left(|\theta_{pred} - \theta_{tgt}|, \pi - |\theta_{pred} - \theta_{tgt}|\right),
\end{equation}
which guarantees that gradients always follow the shortest angular displacement on the circle. Appendix C provides further analysis and discussion on this point.

By simultaneously adopting a periodic, bounded coarse-to-fine update and a shortest-path periodic loss, our decoder refinement for $\theta$ becomes consistent with the orientation setting.  The overall box regression loss concatenates the standard $L_1$ loss for $(c_x,c_y,w,h)$ with $\mathcal{L}_{angle}$ for $\theta$.

\subsection{Oriented Dense O2O}

We introduce \textbf{Oriented Dense O2O} to provide dense supervision while explicitly accelerating the convergence of angle predictions. As shown in Fig. \ref{fig:main2} (E), building upon the Dense O2O \cite{huang2025deim} which combines four replicated images into a single composite grid to increase GT counts, we apply an independent random rotation $\theta_{rot} \in \{0^\circ, 90^\circ, 180^\circ, 270^\circ\}$ to each individual quadrant prior to stitching them together. By artificially enriching the angular diversity within a single training image, this computation-free mechanism forces the model to simultaneously process semantic features at various orientations. Consequently, Oriented Dense O2O seamlessly integrates rotational variance into the dense supervision framework, accelerating the model's convergence on angle predictions and enhancing its robustness against rotated objects.

\section{Experiments}

\subsection{Evaluation Datasets}

\subsubsection{DOTA-1.0.}

DOTA-1.0 \cite{xia2018dota} is a large-scale oriented object detection dataset containing 2,806 images and 188,282 instances across 15 categories, with significant variations in orientation, shape, and scale. Two evaluation protocols are provided: (A) single-scale and (B) multi-scale training and testing. In single-scale settings, images are cropped into $1024 \times 1024$ patches with a stride of 824. In multi-scale settings, images are resized to 0.5, 1.0, and 1.5 scales before being cropped into $1024 \times 1024$ patches with a stride of 524.

\subsubsection{DIOR-R.}

DIOR-R \cite{cheng2022anchor} is constructed from DIOR by adding annotations for rotated bounding boxes. It comprises 23,463 images of size $800 \times 800$, totaling 190,288 oriented instances across 20 categories. The dataset is split into training, validation, and testing sets with a ratio of 1:1:2.

\subsubsection{FAIR-1M-2.0.}

FAIR-1M-2.0 \cite{SUN2022116} is a recently published remote sensing dataset that consists of 15,266 high-resolution images and more than 1 million instances. It contains 5 categories and 37 subcategories. For multi-scale training and testing, images are processed in the same way as in DOTA.

\subsection{Implementation Details}

For DOTA-1.0 and DIOR-R, models are trained on the combined train and val sets and evaluated on the test set. For FAIR-1M-2.0, models are trained on the train set and evaluated on the val set. The results on DOTA are reported using the official evaluation server. The training schedules for different RiO-DETR variants and the corresponding hyperparameter configurations are detailed in Appendix A. All models are trained on a single NVIDIA A800 or H200 GPU. All latencies are measured equally on a single NVIDIA T4 GPU using TensorRT 10 with FP16. For end-to-end models, the latency excludes pre-processing and post-processing time, while for others the \texttt{BatchedNMSPlugin} is employed to get the actual latency under production scenes. Unless otherwise specified, all ablation studies are conducted on the DIOR-R dataset with RiO-DETR-m.

\subsection{Comparison with State-of-the-Art Methods}

\subsubsection{Results on DOTA-1.0.}
Under single-scale training and testing protocol (Table~\ref{table:dota_ss_compare}), RiO-DETR performs strongly across all scales. RiO-DETR-n achieves $78.4\%$ AP$_{50}$ with 2.7 ms latency and 4.0M parameters, surpassing YOLO26n-obb. At the high end, RiO-DETR-x reaches $81.8\%$ AP$_{50}$ at 29.9 ms, outperforming YOLO26x-obb (80.4\%, 30.5 ms) and heavy DETR variants such as RHINO-DETR (79.4\%, 57.0 ms). Under multi-scale evaluation (Table~\ref{tab:dota_ms}), RiO-DETR-x attains $81.76\%$ AP$_{50}$ with the lowest latency among high-precision models. Per-class AP$_{50}$ under multi-scale protocol, results under high precision metrics and results on different backbones are provided in Appendix D.

\subsubsection{Results on DIOR-R.}
RiO-DETR maintains a clear speed–accuracy advantage on DIOR-R (Table~\ref{tab:performance_comparison_dior}). RiO-DETR-s achieves $74.44\%$ AP$_{50}$ at 3.01 ms, while RiO-DETR-x reaches $77.43\%$ at 17.31 ms, exceeding YOLO26x-obb (76.48\%) under similar latency and outperforming prior state-of-the-art models.

\subsubsection{Results on FAIR-1M-2.0.}
On FAIR-1M-2.0 (Table~\ref{tab:performance_comparison_fair1m}), RiO-DETR-x achieves a new state of the art with 47.4 AP$_{50}$ under multi-scale training and testing, surpassing YOLO26x-obb (46.7\%), LSKNet-S (46.3\%), and ReDet (43.2\%), demonstrating strong scalability to large-scale remote sensing benchmarks. Per-class AP$_{50}$ are provided in Appendix D.

\subsection{Efficiency Analysis of RiO-DETR}
Our RiO-DETR series operates at the same latency level as the YOLO26 family across all model scales (n–x). Specifically, RiO-DETR-n/s/m/l/x achieve 2.7/5.2/8.8/13.4/29.9 ms, closely matching the corresponding YOLO26 variants (2.8/4.9/10.2/13.0/30.5 ms), demonstrating that RiO-DETR maintains a fully real-time regime comparable to state-of-the-art CNN-based real-time detectors.

Several CNN-based counterparts \cite{li2025lsknet, yuan2025strip, cai2024poly} that claim efficiency exhibit substantially higher latency (100–200 ms). This gap stems largely from computationally expensive components such as large-kernel convolutions. By explicitly avoiding such high-latency operators, RiO-DETR achieves a clear advantage over these methods. Notably, RiO-DETR is the first oriented DETR to achieve end-to-end real-time inference for a single image, closing the long-standing efficiency gap between transformer-based and CNN-based oriented detectors.

\begin{table*}[t]
\centering

\begin{minipage}{\linewidth}
\centering
\caption{Comparison with state-of-the-art oriented object detectors on DOTA-1.0 under single-scale training and testing protocol.}
\label{table:dota_ss_compare}

\setlength{\tabcolsep}{3pt}
\resizebox{\linewidth}{!}{
\begin{tabular}{ll*{19}{c}}
\toprule
\textbf{Methods} & \textbf{Backbone} & \textbf{\#P} & \textbf{Flops} & \textbf{Lat.} & \textbf{AP$_{50}$} &
PL & BD & BR & GTF & SV & LV & SH & TC & BC & ST & SBF & RA & HA & SP & HC \\
\midrule

\multicolumn{21}{c}{\textbf{CNN-based Non-Real-time Oriented Object Detectors}} \\
\midrule
LSKNet-S \cite{li2025lsknet} & LSKNet-S & 31.0M & 161G & 146.2 & 77.5 & 89.7 & 85.5 & 57.7 & 75.7 & 75.0 & 78.7 & 88.2 & 90.9 & 90.9 & 86.4 & 66.9 & 63.8 & 77.8 & 74.5 & 64.8 \\
PKINet-S \cite{cai2024poly} & PKINet-S & 30.8M & 190G & 187.5 & 78.4 & 89.7 & 84.2 & 55.8 & 77.6 & 80.2 & 84.5 & 88.1 & 90.9 & 87.6 & 86.1 & 66.9 & 70.2 & 77.5 & 73.6 & 62.9 \\
PSD-SQ \cite{feng2024psd} & R-50 & -- & -- & -- & 78.5 & 89.7 & 85.4 & 57.3 & 75.2 & 80.0 & 81.2 & 88.3 & 89.9 & 88.0 & 86.3 & 69.6 & 68.5 & 75.3 & 69.5 & 72.7 \\
RVSA \cite{wang2022advancing} & ViTAE-B & 114.4M & 414G & -- & 79.0 & 89.4 & 84.3 & 59.4 & 73.2 & 80.0 & 85.4 & 88.1 & 90.9 & 88.5 & 86.5 & 58.9 & 72.2 & 77.3 & 79.6 & 71.2 \\
Strip R-CNN-S \cite{yuan2025strip} & StripNet-S & 30.5M & 159G & 106.3 & 80.1 & 88.9 & 86.4 & 57.4 & 76.4 & 79.7 & 84.4 & 88.2 & 90.9 & 86.7 & 87.5 & 69.9 & 66.8 & 79.2 & 82.9 & 75.6 \\
\midrule

\multicolumn{21}{c}{\textbf{DETR-based Non-Real-time Oriented Object Detectors}} \\
\midrule
R. D-DETR~\cite{zeng2024ars} & R-50 
& 41.1M & 409G & 33.6 & 69.5 
& 84.8 & 70.7 & 46.0 & 61.9 & 73.9 & 78.8 & 87.7 & 90.0 & 77.9 & 78.4 
& 47.0 & 54.4 & 66.8 & 67.6 & 55.6 \\
R. D-DETR~\cite{zeng2024ars} & R-50 w/ CSL
& 41.4M & 411G & 33.6 & 72.2 
& 86.2 & 76.6 & 46.6 & 65.2 & 76.8 & 76.3 & 87.7 & 90.7 & 79.3 & 82.3 
& 54.0 & 61.7 & 66.0 & 70.4 & 61.9 \\
EMO2-DETR~\cite{hu2023emo2} & Swin-T 
& -- & -- & -- & 72.3 
& 89.0 & 79.6 & 48.7 & 60.2 & 77.3 & 76.4 & 84.5 & 90.8 & 84.8 & 85.7 
& 48.9 & 67.6 & 66.3 & 71.5 & 53.5 \\
ARS-DETR \cite{zeng2024ars} & R-50 
& 41.4M & 411G & 43.9 & 74.2 
& 86.9 & 75.5 & 48.3 & 69.2 & 77.9 & 77.9 & 87.6 & 90.5 & 77.3 & 82.8
& 60.2 & 64.5 & 74.8 & 71.7 & 66.6 \\
ARS-DETR~\cite{zeng2024ars} & Swin-T 
& 41.9M & 431G & 50.4 & 75.5 
& 87.7 & 76.5 & 50.6 & 69.9 & 79.8 & 83.9 & 87.9 & 90.3 & 86.2 & 85.1 
& 54.6 & 67.0 & 75.6 & 73.7 & 63.4 \\
AO2-DETR \cite{dai2022ao2} & R-50 & 74.3M & 304G & -- & 77.7 & 89.3 & 85.0 & 56.7 & 74.9 & 78.9 & 82.7 & 87.3 & 90.5 & 84.7 & 85.4 & 62.0 & 70.0 & 74.7 & 72.4 & 71.6 \\
RHINO-DETR \cite{lee2025hausdorff} & R-50 
& 47.6M & 566G & 47.8 & 78.7 
& 88.2 & 85.1 & 55.8 & 72.7 & 80.2 & 83.1 & 89.0 & 90.8 & 87.1 & 86.8 
& 65.3 & 71.6 & 77.7 & 81.2 & 64.7 \\
RHINO-DETR \cite{lee2025hausdorff} & Swin-T & 50.8M & 609G & 57.0 & 79.4 & 88.2 & 84.8 & 58.5 & 77.7 & 81.1 & 85.6 & 89.2 & 90.9 & 87.0 & 86.4 & 65.5 & 71.3 & 78.2 & 82.8 & 64.3 \\
Oriented-DETR \cite{zhao2024projecting} & R-50 
& 57.2M & 302G & 87.2 & 79.1 
& 89.2 & 86.4 & 57.7 & 75.3 & 81.1 & 84.7 & 89.1 & 90.9 & 86.1 & 87.0 
& 59.5 & 70.3 & 79.3 & 81.5 & 68.8 \\
Oriented-DETR \cite{zhao2024projecting}  & Swin-T 
& 57.7M & 309G & 89.8 & 79.8 
& 89.4 & 85.1 & 57.8 & 75.0 & 81.2 & 86.1 & 89.1 & 90.9 & 88.7 & 87.0 
& 62.9 & 69.1 & 80.7 & 82.8 & 71.0 \\
\midrule
\multicolumn{21}{c}{\textbf{Real-time Oriented Object Detectors}} \\
\midrule
PPYOLOE-R-l \cite{xu2022pp} & CRN-l & 52.2M & 141G & 20.5 & 78.1 & 89.2 & 81.0 & 54.0 & 70.2 & 81.8 & 85.2 & 88.8 & 90.8 & 87.0 & 88.0 & 62.9 & 67.9 & 76.6 & 79.1 & 69.7 \\
PPYOLOE-R-x \cite{xu2022pp} & CRN-x & 98.4M & 264G & 32.8 & 78.3 & 89.5 & 79.7 & 55.0 & 75.6 & 82.4 & 85.2 & 88.3 & 90.8 & 85.7 & 87.7 & 63.2 & 69.5 & 77.1 & 75.1 & 69.4 \\
RTMDet-R-m \cite{lyu2022rtmdet} & CSPNext-m & 24.7M & 100G & 14.9 & 78.2 & 89.2 & 84.7 & 53.9 & 74.7 & 81.5 & 84.0 & 88.7 & 90.8 & 87.4 & 87.2 & 59.4 & 66.7 & 77.7 & 82.4 & 65.3 \\
RTMDet-R-l \cite{lyu2022rtmdet} & CSPNext-l & 52.3M & 205G & 23.7 & 78.8 & 89.4 & 84.2 & 55.2 & 75.1 & 80.8 & 84.5 & 89.0 & 90.9 & 87.4 & 87.2 & 63.1 & 67.9 & 78.1 & 80.8 & 69.1 \\
YOLO26n-obb \cite{sapkota2025yolo26} & YOLO26n & 2.5M & 14G & \underline{2.8} & \underline{77.7} & 89.7 & 84.7 & 52.1 & 71.1 & 83.1 & 78.9 & 88.7 & 91.1 & 87.3 & 86.9 & 58.5 & 69.1 & 76.6 & 82.8 & 65.1 \\
YOLO26s-obb \cite{sapkota2025yolo26} & YOLO26s & 9.8M & 55G & \textbf{4.9} & \underline{79.7} & 89.7 & 86.3 & 55.5 & 71.8 & 82.5 & 81.9 & 88.9 & 91.1 & 88.2 & 88.1 & 62.9 & 72.5 & 78.0 & 83.9 & 74.5 \\
YOLO26m-obb \cite{sapkota2025yolo26} & YOLO26m & 21.2M & 183G & \underline{10.2} & \underline{80.0} & 89.2 & 86.8 & 56.4 & 73.0 & 82.6 & 78.3 & 89.1 & 91.2 & 88.3 & 87.9 & 64.5 & 78.0 & 78.3 & 83.1 & 79.7 \\
YOLO26l-obb \cite{sapkota2025yolo26} & YOLO26l & 25.6M & 230G & \textbf{13.0} & \underline{80.2} & 89.5 & 86.3 & 56.2 & 74.2 & 82.8 & 78.8 & 89.3 & 91.1 & 88.6 & 88.1 & 67.8 & 72.2 & 78.2 & 83.5 & 76.5 \\
YOLO26x-obb \cite{sapkota2025yolo26} & YOLO26x & 57.6M & 517G & \underline{30.5} & \underline{80.4} & 89.7 & 86.6 & 57.3 & 73.8 & 82.8 & 79.2 & 89.2 & 91.1 & 89.1 & 88.1 & 66.7 & 70.6 & 77.9 & 83.3 & 80.7 \\
\midrule
\textbf{RiO-DETR-n} & HGNet-B0 & 4.0M & 17G & \textbf{2.7} & \textbf{78.4} & 87.3 & 86.1 & 54.9 & 73.6 & 80.5 & 85.1 & 88.0 & 90.8 & 87.4 & 86.9 & 60.9 & 72.0 & 76.8 & 73.4 & 71.8 \\
\textbf{RiO-DETR-s} & HGNet-B0 & 8.2M & 53G & \underline{5.2} & \textbf{80.3} & 85.6 & 84.8 & 57.5 & 75.5 & 80.7 & 86.2 & 89.0 & 90.8 & 88.4 & 88.2 & 65.2 & 74.2 & 78.3 & 80.1 & 79.6 \\
\textbf{RiO-DETR-m} & HGNet-B2 & 18.6M & 158G & \textbf{8.8} & \textbf{80.9} & 85.9 & 86.4 & 61.7 & 78.1 & 82.0 & 86.8 & 89.2 & 90.8 & 88.9 & 87.7 & 67.4 & 73.3 & 78.8 & 77.4 & 78.8 \\
\textbf{RiO-DETR-l} & HGNet-B4 & 27.5M & 230G & \underline{13.4} & \textbf{81.7} & 86.7 & 86.8 & 60.8 & 79.4 & 82.0 & 86.4 & 89.0 & 90.8 & 88.7 & 88.5 & 70.0 & 76.0 & 79.2 & 83.4 & 76.9 \\
\textbf{RiO-DETR-x} & HGNet-B5 & 62.5M & 527G & \textbf{29.9} & \textbf{81.8} & 88.3 & 87.5 & 61.9 & 79.2 & 83.2 & 86.5 & 89.2 & 90.8 & 88.4 & 87.8 & 71.2 & 75.1 & 78.5 & 82.2 & 77.1 \\
\bottomrule
\end{tabular}
}
\end{minipage}

\vspace{12pt}

\begin{minipage}{\linewidth}
\centering
\caption{Comparison with state-of-the-art oriented object detectors on DOTA-1.0 under multi-scale training and testing protocol.}
\label{tab:dota_ms}

\setlength{\tabcolsep}{4pt}
\renewcommand{\arraystretch}{1.2}
\resizebox{\linewidth}{!}{
\begin{tabular}{lccccccccc}
\toprule
\textbf{Model} & 
\makecell{PKINet-S \\ \cite{cai2024poly}} & 
\makecell{LSKNet-S \\ \cite{li2025lsknet}} & 
\makecell{Strip R-CNN-S \\ \cite{yuan2025strip}} & 
\makecell{PPYOLOE-R-x \\ \cite{xu2022pp}} & 
\makecell{RTMDet-R-l \\ \cite{lyu2022rtmdet}} & 
\makecell{YOLO26m-obb \\ \cite{sapkota2025yolo26}} & 
\textbf{Ours-m} & 
\makecell{YOLO26x-obb \\ \cite{sapkota2025yolo26}} & 
\textbf{Ours-x} \\
\midrule
\textbf{Latency} & 187.5 & 146.2 & 106.3 & 32.8 & 23.7 & 10.2 & \textbf{8.8} & 30.5 & \textbf{29.9} \\
\textbf{AP$_{50}$}  &  81.06  & 81.64 & 82.28 & 80.73 & 81.33 & 81.00  & \textbf{81.49} & 81.70 & \textbf{81.76} \\
\bottomrule
\end{tabular}
}
\end{minipage}

\end{table*}

\begin{table*}[t]
\centering
\setlength{\tabcolsep}{3pt} 
\renewcommand{\arraystretch}{1.05}
\footnotesize

\begin{minipage}[t]{0.48\textwidth}
\centering
\captionof{table}{Performance comparison with state-of-the-art oriented object detectors on DIOR-R.}
\label{tab:performance_comparison_dior}

\resizebox{\linewidth}{!}{%
\begin{tabular}{llcccc}
\toprule
\textbf{Methods} & \textbf{Backbone} & \textbf{\#P} & \textbf{Flops} & \textbf{Lat.} & \textbf{AP$_{50}$} \\
\midrule
\multicolumn{6}{c}{\textbf{Non-Real-time Oriented Object Detectors}} \\ \midrule
LSKNet-S \cite{li2025lsknet} & LSKNet-S & 31.0M & 111G & 123.62 & 65.90 \\
PKINet-S \cite{cai2024poly} & PKINet-S & 30.8M & 118G & 149.91 & 67.03 \\
Strip R-CNN-S \cite{yuan2025strip} & StripNet-S & 30.5M & 157G & 116.75 & 68.70 \\
RHINO-DETR \cite{lee2025hausdorff} & Swin-T & 50.8M & 383G & 35.56 & 72.67 \\
\midrule
\multicolumn{6}{c}{\textbf{Real-time Oriented Object Detectors}} \\ \midrule
YOLO26n-obb \cite{sapkota2025yolo26} & YOLO26n & 2.5M & 10G & \underline{2.23} & \underline{71.88} \\
YOLO26s-obb \cite{sapkota2025yolo26} & YOLO26s & 9.8M & 39G & \textbf{2.84} & \underline{74.07} \\
YOLO26m-obb \cite{sapkota2025yolo26} & YOLO26m & 21.2M & 129G & \underline{5.91} & \underline{74.65} \\
YOLO26l-obb \cite{sapkota2025yolo26} & YOLO26l & 25.6M & 158G & \textbf{7.53} & \underline{75.31} \\
YOLO26x-obb \cite{sapkota2025yolo26} & YOLO26x & 57.6M & 353G & \underline{17.66} & \underline{76.48} \\
\midrule
\textbf{RiO-DETR-n} & HGNet-B0 & 4.0M & 11G & \textbf{2.20} & \textbf{71.92} \\
\textbf{RiO-DETR-s} & HGNet-B0 & 8.2M & 16G & \underline{3.01} & \textbf{74.44} \\
\textbf{RiO-DETR-m} & HGNet-B2 & 18.6M & 97G & \textbf{5.10} & \textbf{75.73} \\
\textbf{RiO-DETR-l} & HGNet-B4 & 27.5M & 141G & \underline{7.76} & \textbf{76.11} \\
\textbf{RiO-DETR-x} & HGNet-B5 & 62.7M & 324G & \textbf{17.31} & \textbf{77.43} \\
\bottomrule
\end{tabular}%
}
\end{minipage}
\hfill
\begin{minipage}[t]{0.48\textwidth}
\centering
\captionof{table}{Performance comparison with SoTA methods on FAIR-1M-2.0 under multi-scale training and testing protocol.}
\label{tab:performance_comparison_fair1m}

\resizebox{\linewidth}{!}{%
\begin{tabular}{lccccc}
\toprule
\textbf{Methods} & \textbf{Backbone} & \textbf{\#P} & \textbf{Flops} & \textbf{Lat.} & \textbf{AP$_{50}$} \\
\midrule
SASM RepPoints \cite{hou2022shape} & R-101 & 55.8M & 542G & 101.9 & 30.9 \\
R-FCOS \cite{tian2019fcos} & R-101 & 50.9M & 284G & 44.8 & 36.1 \\
S2A-Net \cite{yujie2024s2anet} & R-50 & 31.6M & 588G & -- & 37.4 \\
R-Faster RCNN \cite{yang2020rotated} & R-101 & 60.1M & 289G & 67.8 & 37.5 \\
O-RepPoints \cite{li2022oriented} & Swin-T & 37.3M & 200G & -- & 38.9 \\
O-RCNN \cite{xie2021oriented} & R-101 & 60.3M & 289G & 119.0 & 40.4 \\
RoI Trans. \cite{ding2019learning} & R-101 & 67.8M & 607G & 84.5 & 40.2 \\
LOOD (RT) \cite{zhou2024boosting} & R-50 & -- & -- & -- & 42.6 \\
ReDet \cite{han2021redet} & ReR-50 & 31.8M & 225G & 81.3 & 43.2 \\
PKINet-S \cite{cai2024poly} & PKINet-S & 30.8M  & 190G & 187.5 & 44.5 \\
LOOD (RD) \cite{zhou2024boosting} & R-101 & -- & -- & -- & 44.9 \\
Strip-RCNN-S \cite{yuan2025strip} & StripNet-S & 30.5M & 159G & 127.1 & 45.3 \\
LSKNet-S \cite{li2025lsknet} & LSKNet-S & 31.0M & 161G & 146.2 & 45.8 \\
\midrule
YOLO26m-obb \cite{sapkota2025yolo26} & YOLO26m & 21.2M & 183G & \underline{10.2} & \underline{42.5} \\
YOLO26x-obb \cite{sapkota2025yolo26} & YOLO26x & 57.6M & 517G & \underline{30.5} & \underline{46.7} \\
\midrule
\textbf{RiO-DETR-m} & HGNet-B2 & 18.6M & 158G & \textbf{8.8} & \textbf{43.6} \\
\textbf{RiO-DETR-x} & HGNet-B5 & 62.5M & 527G & \textbf{29.9} & \textbf{47.4} \\
\bottomrule
\end{tabular}%
}
\end{minipage}

\end{table*}

\subsection{Ablation Studies}

\begin{table}[t]
\centering
\scriptsize
\caption{Step-by-step modifications from baseline model to RiO-DETR-m on DIOR-R. Each step shows changes in AP$_{50}$, the number of parameters, latency, and Flops.}
\label{tab:ablation_study}
\begin{tabular}{lcccc}
\toprule
\textbf{Methods} & \textbf{\#P} & \textbf{Flops} & \textbf{Latency} & \textbf{AP$_{50}$} \\ 
\midrule
Oriented RT-DETRv2 \textbf{(Ours Scratch Implementation)} & 18.61M & 97.06G & 5.04 & 70.35 \\
+ Hausdorff Matching from RHINO-DETR \cite{lee2025hausdorff} \& KLD Loss \cite{yang2021learning} &  &  & & 72.86 \\
+ Universal Matching Strategy from D-FINE \cite{peng2024d} &  &  & & 73.33 \\
+ Dense O2O from DEIM \cite{huang2025deim} (\textbf{Our Implemented Baseline}) & 18.61M & 97.06G & 5.04 & 73.47 \\
\midrule
\textit{+ Content-Driven Angle Estimation} & & & & \\
\quad \textit{Geometry-Decoupled Query Encoding} & 18.59M & 97.01G & 5.04 & 74.18 \\
\quad \textit{Rotation-Rectified Orthogonal Attention} & 18.67M & 97.18G & 5.10 & 74.74 \\
\textit{+ Decoupled Periodic Refinement} &  &  & & 75.46 \\
\textit{+ Oriented Dense O2O} (\textbf{Our RiO-DETR-m}) & 18.67M & 97.18G & 5.10 & 75.73 \\
\bottomrule
\end{tabular}
\end{table}

\subsubsection{The Roadmap to RiO-DETR.}

Table \ref{tab:ablation_study} showcases the stepwise progression from the baseline model to our proposed RiO-DETR framework. Starting from RT-DETRv2 adding a simple OBB head, we first introduce a stronger bipartite matching and optimization objective, which yields a notable improvement in AP$_{50}$. Specifically, we formulate the matching cost $\mathcal{C}_{match}$ as a combination of focal loss, KLD loss, and Hausdorff distance:
$$\mathcal{C}_{match} = \lambda_{cls} \mathcal{C}_{focal} + \lambda_{kld} \mathcal{C}_{kld} + \lambda_{hausdorff} \mathcal{C}_{hausdorff}$$
Correspondingly, the overall training loss $\mathcal{L}$ is optimized using a shortest-path periodic L1 loss ($\mathcal{L}_{angle}$) alongside the focal and KLD losses:
$$\mathcal{L} = \lambda_{cls} \mathcal{L}_{focal} + \lambda_{kld} \mathcal{L}_{kld} + \lambda_{angle} \mathcal{L}_{angle}$$
By integrating the matching strategy from D-FINE and the training scheme from DEIM, the baseline increases to 73.47 AP$_{50}$. On this basis, Content-Driven Angle Estimation yields steady gains: Geometry-Decoupled Query Encoding raises AP$_{50}$ to 74.18 with negligible cost, and Rotation-Rectified Orthogonal Attention further improves it to 74.74 with slight latency overhead. Finally, Decoupled Periodic Refinement and Oriented Dense O2O push performance to 75.73 AP$_{50}$ while keeping parameters, FLOPs, and latency nearly unchanged.

\begin{table*}[t]
\centering
\tiny 

\begin{minipage}[t]{0.48\linewidth} 
    \centering
    \caption{Comparison of our implemented baseline with SoTA methods on DIOR-R.}
    \label{tab:baseline_comparison}
    \resizebox{\linewidth}{!}{
        \begin{tabular}{lcccc}
        \toprule
        \textbf{Methods} & \textbf{\#P} & \textbf{Flops} & \textbf{Latency} & \textbf{AP$_{50}$} \\
        \midrule
        RoI Trans. \cite{ding2019learning} & 68.1M & 373G & -- & 63.87 \\
        RVSA-ViTAE-B \cite{wang2022advancing} & 114.4M & -- & -- & 71.05 \\
        DCFL-ReR-101 \cite{xu2023dynamic} & -- & -- & -- & 71.03 \\
        LSKNet-S \cite{li2025lsknet} & 31M & 111G & 123.62 & 65.90 \\
        PKINet-S \cite{cai2024poly} & 30.8M & 118G & 149.91 & 67.03 \\
        Strip R-CNN-S \cite{yuan2025strip} & 30.5M & 157G & 116.75 & 68.70 \\
        RHINO-DETR-Swin-T \cite{lee2025hausdorff} & 50.8M & 383G & 35.56 & 72.67 \\
        YOLO26m-obb \cite{sapkota2025yolo26} & 21.2M & 129G & \underline{5.91} & \textbf{74.65} \\
        \midrule
        \textbf{Baseline} & 18.61M & 97G & \textbf{5.04} & \underline{73.47} \\
        \bottomrule
        \end{tabular}
    }
\end{minipage}
\hfill
\begin{minipage}[t]{0.48\linewidth}
    \centering
    \caption{Ablation on Oriented Dense O2O. Epoch notes the one with highest AP$_{50}$ on Stage 1.}
    \label{tab:o2o_ablation}
    \begin{tabularx}{\linewidth}{l *{2}{>{\centering\arraybackslash}X}}
    \toprule
    \textbf{Methods} & \textbf{Epoch} & \textbf{AP$_{50}$} \\
    \midrule
    No Augmentation & 94 & 72.18 \\
    Regular Augmentation & 86 & 73.33 \\
    Dense O2O \cite{huang2025deim} & 68 & 73.47 \\
    \textbf{Oriented Dense O2O (Ours)} & \textbf{60} & \textbf{73.88} \\
    Dense O2O w/ random rotation & 62 & 73.66 \\
    Pure Angle Augmentation & 84 & 73.53 \\
    \bottomrule
    \end{tabularx}
\end{minipage}

\vspace{2em} 

\begin{minipage}[t]{0.48\linewidth}
    \centering
    \caption{Ablation study on the components of Decoupled Periodic Refinement.}
    \label{tab:dpr_components}
    \begin{tabularx}{\linewidth}{*{3}{>{\centering\arraybackslash}X}}
    \toprule
    \textbf{SP-$L_1$} & \textbf{Periodic} & \textbf{AP$_{50}$} \\
    \midrule
              &            & 74.18 \\
    \checkmark &            & 74.32 \\
              & \checkmark & 74.05 \\
    \checkmark & \checkmark & \textbf{74.74} \\
    \bottomrule
    \end{tabularx}
\end{minipage}
\hfill
\begin{minipage}[t]{0.48\linewidth}
    \centering
    \caption{Ablation study on Geometry-Decoupled Query Encoding.}
    \label{tab:pe_masking}
    \begin{tabularx}{\linewidth}{*{4}{>{\centering\arraybackslash}X}}
    \toprule
    \textbf{Center} & \textbf{Size} & \textbf{Angle} & \textbf{AP$_{50}$} \\
    $(cx, cy)$ & $(w, h)$ & $(\theta)$ & \\
    \midrule
    \checkmark & & & 72.56 \\
    \checkmark & & \checkmark & 72.34 \\
    \checkmark & \checkmark & \checkmark & 73.57 \\
    \checkmark & \checkmark & & \textbf{73.81} \\
    \bottomrule
    \end{tabularx}
\end{minipage}
\end{table*}

\begin{table}[t]
\begin{minipage}[b]{1.0\linewidth}
    \centering
    \tiny 
    \setlength{\tabcolsep}{12pt}
    \caption{Ablation study on Attention Head Splitting Strategy for Rotation-Rectified Orthogonal Attention. H denotes the number of attention heads (default H = 8)}
    \label{tab:head_splitting}
    \begin{tabular}{lcc}
    \toprule
    \textbf{Strategy} & \textbf{Ratio} ($\theta : \theta + \pi/2$) & \textbf{AP$_{50}$} \\
    \midrule
    Vanilla Alignment & $8 : 0$ & 73.81 \\
    Asymmetric Alignment & $6 : 2$ & 74.02 \\
    Multi-angle Distribution & $2:2:2:2^\ast$ & 73.95 \\
    \textbf{Symmetric Orthogonal} & $4 : 4$ & \textbf{74.18} \\
    \bottomrule
    \addlinespace[2pt]
    \multicolumn{3}{l}{\scriptsize $^\ast$ Multi-angle includes $\theta, \theta + \pi/4, \theta + \pi/2$, and $\theta + 3\pi/4$.} \\
    \end{tabular}
\end{minipage}
\end{table}

\subsubsection{Ablation on Geometry-Decoupled Query Encoding.}

Table~\ref{tab:pe_masking} analyzes different geometry components used in query encoding. Encoding the center coordinates alone achieves limited performance (72.56 AP$_{50}$), and additionally encoding the angle further degrades accuracy (72.34), suggesting that directly injecting $\theta$ into the query can introduce optimization ambiguity. Incorporating size information provides a clear gain (73.57), and the best result is obtained by encoding center and size while leaving the angle to be inferred from content features, reaching 73.81 AP$_{50}$. These results validate our design choice of geometry decoupling: use $(cx,cy,w,h)$ to guide localization while predicting $\theta$ in a content-driven manner.

\subsubsection{Ablation on Decoupled Periodic Refinement Components.}


Table \ref{tab:dpr_components} analyzes the contribution of each component in Decoupled Periodic Refinement. Using SP-$L_1$ alone slightly improves AP$_{50}$ from 74.18 to 74.32. The periodic update mechanism alone leads to a small drop in AP${50}$ (74.18 $\rightarrow$ 74.05), suggesting that constraining angles within $[0,\pi)$ without a consistent periodic loss is insufficient and may introduce instability. Combining both components yields the largest gain, reaching 74.74 AP$_{50}$, demonstrating that consistent periodic modeling in both the loss and update steps is crucial for stable OBB refinement.

\subsubsection{Ablation on Oriented Dense O2O.}

Compared to Regular Augmentation (73.33 AP$_{50}$ at 86 epochs), Dense O2O achieves 73.47 AP$_{50}$ in 68 epochs, proving that denser supervision mitigates sparse matching. Oriented Dense O2O reaches the highest AP$_{50}$ (73.88) with the fastest convergence, demonstrating the value of rotation diversity. While random rotation yields a smaller gain (73.66), we attribute this to noisy angle supervision from orientation ambiguity in near-square objects. In contrast, the discrete rotations in Oriented Dense O2O enrich patterns while reducing ambiguity, resulting in more stable training and superior performance. Oriented Dense O2O is essentially equivalent to performing angle-stratified dense supervision within a single composite image, which improves the angular separability of targets in each Hungarian matching step. Moreover, using discrete rotations maximizes angular coverage while minimizing equivalent-parameterization noise of near-square objects; as a result, the angle branch enters a stable refinement regime earlier, with lower gradient variance and reduced jitter near periodic boundaries. Further analysis is provided in Appendix E.

\subsubsection{Ablation on the Design of Rotation-Rectified Orthogonal Attention.}

Table~\ref{tab:head_splitting} analyzes head splitting in Rotation-Rectified Orthogonal Attention. Aligning all heads to $\theta$ yields 73.81 AP$_{50}$, while an asymmetric $6{:}2$ split slightly improves performance to 74.02. The symmetric orthogonal design ($4{:}4$ for $\theta$ and $\theta+\pi/2$) performs best at 74.18, showing that modeling orthogonal directions better captures longitudinal and lateral structures. Distributing heads across four angles further reduces performance (73.95), due to diluted capacity per direction.

\begin{figure}[t]
  \centering
  \includegraphics[height=0.15\textheight, width=\linewidth, keepaspectratio]{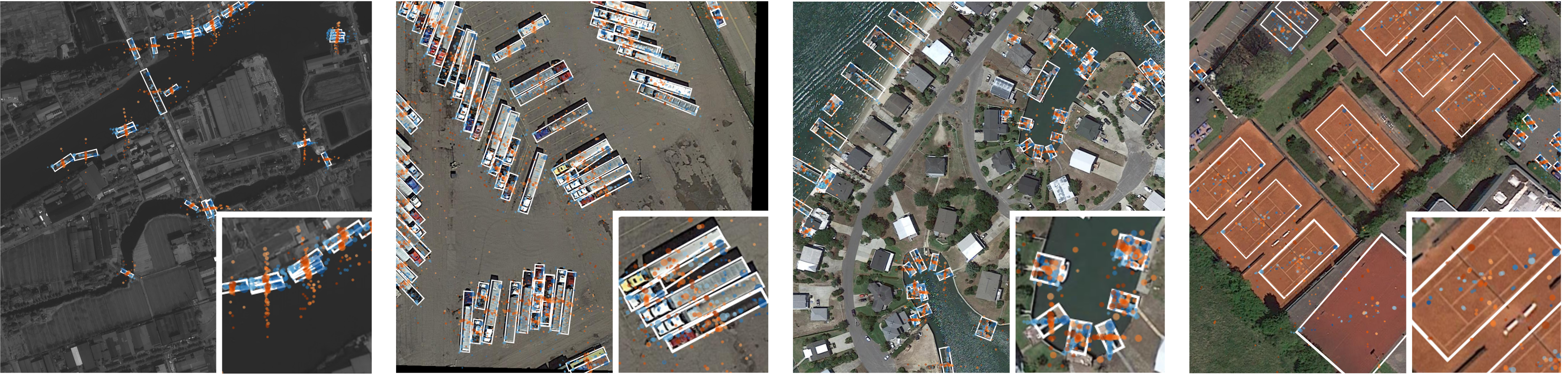}
  \caption{Visualization of deformable attention sampling points on DOTA.}
  \label{fig:attention_vis}
\end{figure}

\begin{figure}[t]
  \centering
  
  \begin{minipage}{0.48\textwidth}
    \centering
    \includegraphics[width=\linewidth]{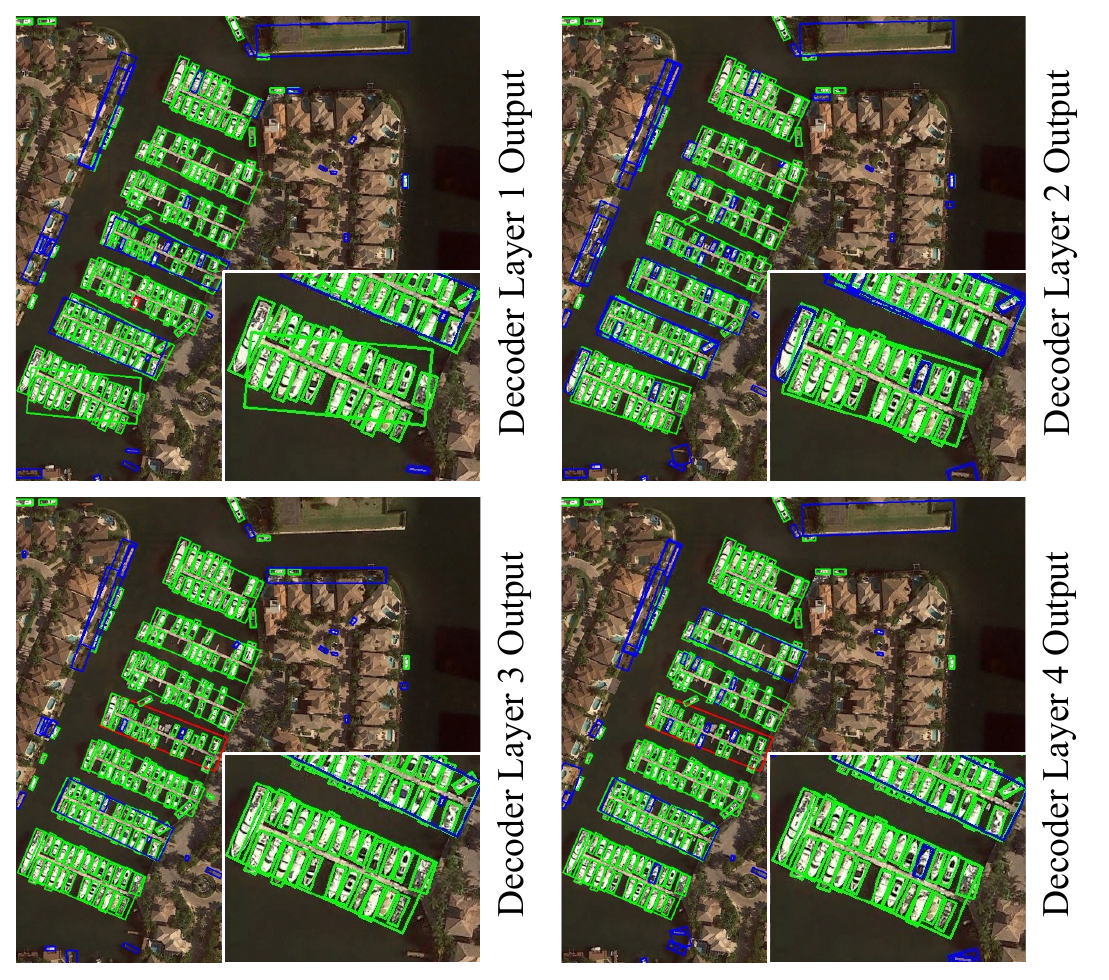}
    \caption{Visual illustration of layer-wise angular refinement for a specific instance.}
    \label{fig:layerwise_vis}
  \end{minipage}
  \hfill 
  \begin{minipage}{0.48\textwidth}
    \centering
    \includegraphics[width=\linewidth]{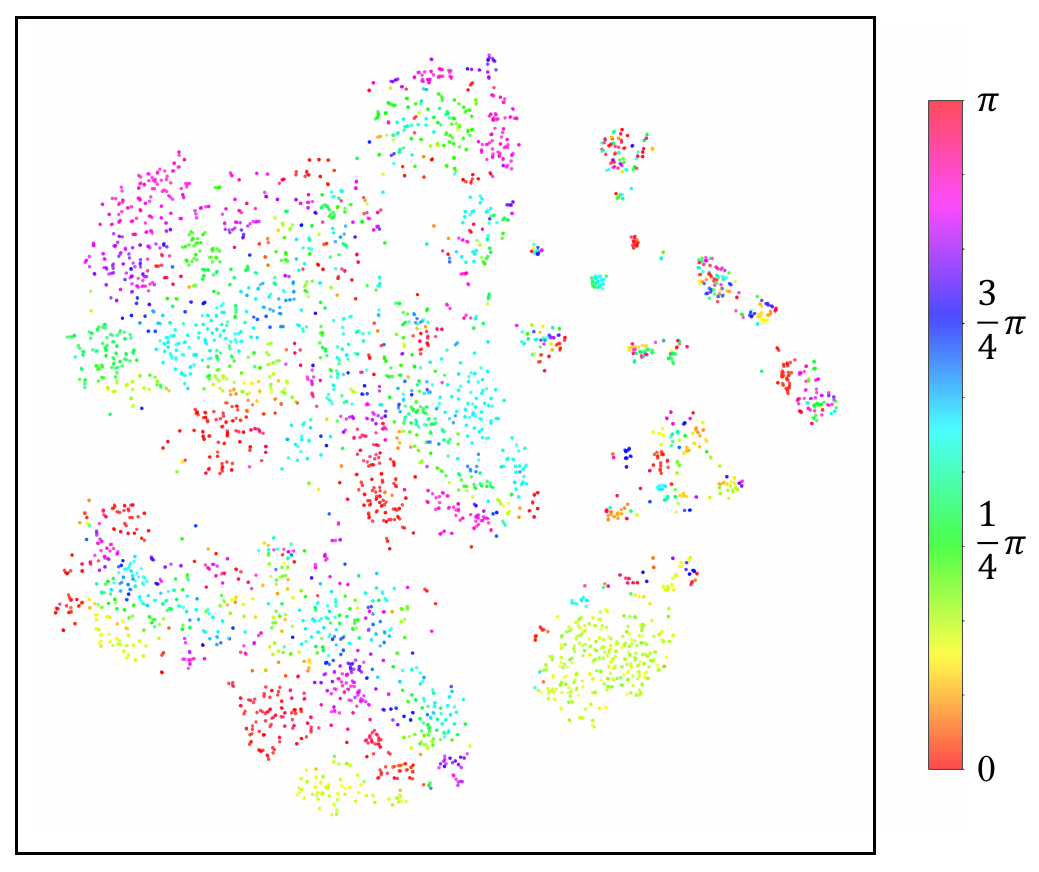}
    \caption{t-SNE visualization of angular features, demonstrating Geometry-Decoupled Query Encoding's clustering.}
    \label{fig:angle_t_sne}
  \end{minipage}
\end{figure}

\subsection{Visualization Analysis}

\subsubsection{Visualization Analysis of Sampling Patterns.}

To intuitively understand how Rotation-Rectified Orthogonal Attention improves feature extraction, we visualize the deformable attention sampling points in Fig. \ref{fig:attention_vis}. Our proposed Rotation-Rectified Orthogonal Attention enforces an orthogonal sampling pattern, effectively capturing both longitudinal and lateral structural features of the oriented objects.

\subsubsection{Visualization Analysis of Angular Prediction.}

Figure \ref{fig:layerwise_vis} illustrates this refinement on a challenging instance. Starting from a coarse estimate, the predicted OBB progressively aligns with object boundaries: Layer 1 corrects the global heading, while Layers 2–4 remove minor angular deviations without overshooting, further validating the stability and precision of the proposed strategy. More cases are provided in Appendix E.

\subsubsection{Visualizing Angular Feature Distribution.}

To further validate the effectiveness of Geometry-Decoupled Query Encoding, we employ t-SNE to visualize the query angular features on FAIR-1M-2.0 in Fig. \ref{fig:angle_t_sne}. Our decoupled formulation yields content-query embeddings that organize into clear and separated clusters across orientation intervals, indicating that the model encodes orientation-sensitive semantic evidence in the content space rather than relying on a rigid geometric prior, which is consistent with our design choice to avoid injecting $\theta$ as positional geometry. Further t-SNE analyses are provided in Appendix E.

\section{Conclusion}
In this paper, we present \textbf{RiO-DETR}, the first real-time oriented detection transformer that offers a meaningful step forward in oriented object detection. Content-Driven Angle Estimation, Decoupled Periodic Refinement, and Oriented Dense O2O jointly address angle-content entanglement, periodicity-induced optimization instability, and slow convergence of existing methods. Extensive experiments on DOTA-1.0, DIOR-R, and FAIR-1M-2.0 demonstrate a new accuracy–latency frontier compared to all state-of-the-art CNN and DETR counterparts. \textbf{Limitation and Future Work:} While RiO-DETR significantly optimizes the detection pipeline, designing a \emph{real-time} yet high-performance backbone specifically tailored for the complexities of oriented/remote sensing object detection remains an open challenge. Future work may focus on specialized feature extraction to further enhance the real-time trade-off. We hope RiO-DETR serves as a strong, practical baseline for the community and inspires further exploration into real-time end-to-end perception for oriented objects. 

\section*{Acknowledgements}
This work was in part supported by the National Natural Science Foundation of China under grants 62472399 and Open Fund of APKL of BIIP, IAI, Hefei Comprehensive National Science Center under grants 24YGXT003.

%
%
\bibliographystyle{splncs04}
\bibliography{main}
\end{document}


\title{\emph{Appendix of} RiO-DETR: DETR for Real-time Oriented Object Detection}

\author{}
\authorrunning{}
\institute{}

\maketitle

\section{More Implementation Details}

\subsection{More Details about Our Implemented Baseline}

As described in Sec.~3.1, because the community lacked real-time DETR-style oriented detectors, we built a baseline by extending the official RT-DETRv2 codebase with full oriented-detection support~\cite{lv2024rt}.

Specifically, we added an OBB data pipeline (loading, format conversion, and geometry-aware augmentation), an oriented regression head that outputs a 5D box parameterization, as well as the corresponding post-processing utilities and submission interface required by oriented benchmarks. Some classes and functions were adapted from MMRotate for implementation convenience \cite{zhou2022mmrotate}. To make the baseline stable and competitive under one-to-one matching, we followed RHINO-DETR’s rotated-DETR practice by replacing the IoU-style regression objective with a KLD-based loss, and further adopted the Hausdorff-distance matching cost to reduce assignment ambiguity caused by angle periodicity and near-square instances~\cite{lee2025hausdorff}. Building on recent advances in real-time DETRs, we further incorporated (i) the universal matching strategy from D-FINE to improve matching robustness and optimization behavior~\cite{peng2024d}, and (ii) the Dense O2O training scheme from DEIM to densify supervision without adding extra heads or decoders, thus accelerating convergence while preserving end-to-end inference~\cite{huang2025deim}.

These modifications produced an efficient real-time oriented DETR baseline, but did not resolve the core challenges of DETR-based oriented object detection. It provided a strong foundation for our subsequent task-native designs.

\subsection{More Implementation Details of RiO-DETR Series}

Table \ref{tab:trod_hyperparameters} summarizes the hyperparameter settings of the RiO-DETR models. All variants adopt HGNetV2 backbones pretrained on ImageNet\cite{deng2009imagenet} and use the AdamW optimizer. To support different computational budgets, the model scale is adjusted by reducing the embedding dimension from 384 to 128 and decreasing the number of decoder layers from 4 in larger models to 3 in lighter ones. Most variants use three feature levels, while RiO-DETR-n adopts two levels for higher efficiency. Across all models, the number of queries is fixed at 300, and the same loss weights are used, with both $\mathcal{L}_{\text{L1}}$ and $\mathcal{L}_{\text{KLD}}$ set to 5. We follow training strategies similar to RT-DETRv2. In particular, the training schedule is scaled inversely with model size: larger models such as RiO-DETR-x are trained for 72 epochs with a batch size of 16, whereas the Nano variant is trained for 160 epochs with a batch size of 128.

\begin{table}[!htbp]
\caption{Hyperparameter configurations for RiO-DETR models on DOTA-1.0.}
\vspace{-0.4cm}
\begin{center}
\resizebox{\linewidth}{!}{
\begin{tabular}{l|ccccc}
\toprule
\textbf{Setting} & \textbf{RiO-DETR-x} & \textbf{RiO-DETR-l} & \textbf{RiO-DETR-m} & \textbf{RiO-DETR-s} & \textbf{RiO-DETR-n} \\
\midrule
Backbone Name & HGNetv2-B5 & HGNetv2-B4 & HGNetv2-B2 & HGNetv2-B0 & HGNetv2-B0 \\
Optimizer & AdamW & AdamW & AdamW & AdamW & AdamW \\
Embedding Dimension & 384 & 256 & 256 & 224 & 128 \\
Feedforward Dimension & 2048 & 1024 & 1024 & 1024 & 512 \\
Decoder Layers & 4 & 4 & 3 & 3 & 3 \\
Queries & 300 & 300 & 300 & 300 & 300 \\
Number of Levels & 3 & 3 & 3 & 3 & 2 \\
\midrule
Base LR & 8e-5 & 8e-5 & 1e-4 & 1e-4 & 2e-4 \\
Backbone LR & 8e-6 & 8e-6 & 1e-5 & 1e-5 & 2e-5 \\
Weight Decay & 1.25e-4 & 1.25e-4 & 1e-4 & 1e-4 & 1e-4 \\
\midrule
Weight of $\mathcal{L}_{\text{Focal}}$ & 1 & 1 & 1 & 1 & 1 \\
Weight of $\mathcal{L}_{\text{L1}}$ & 5 & 5 & 5 & 5 & 5 \\
Weight of $\mathcal{L}_{\text{KLD}}$ & 5 & 5 & 5 & 5 & 5 \\
\midrule
Total Batch Size & 16 & 16 & 16 & 16 & 128 \\
Total Epochs & 72 & 102 & 102 & 132 & 160 \\
Flat Epochs & 20 & 51 & 51 & 66 & 80 \\
No Aug Epochs & 12 & 12 & 12 & 12 & 12 \\
\bottomrule
\end{tabular}
}
\vspace{-0.5cm}
\end{center}
\label{tab:trod_hyperparameters}
\end{table}

\subsection{Details of Latency Measurement}

For fairness, all models are evaluated for end-to-end inference latency under a single NVIDIA T4 GPU with TensorRT.

\textbf{Engine execution.} Each model is serialized into a TensorRT engine and executed via \texttt{IExecutionContext}. For TensorRT 10 engines, inference is launched with \texttt{execute\_async\_v3}; otherwise we fall back to \texttt{execute\_async\_v2}.

\textbf{I/O shapes and memory binding.} For dynamic-shape engines, we choose optimization profile 0 and benchmark at the OPT shape. GPU buffers for all inputs/outputs are allocated once as CUDA tensors, and their device pointers are bound to TensorRT using tensor addresses (TensorRT 10) or bindings (legacy). This avoids repeated allocation overhead and ensures the benchmark measures pure GPU inference.

\textbf{Warm-up.} We run 50 warm-up iterations before timing to stabilize GPU states and eliminate one-time overhead (e.g., CUDA context initialization). All warm-up iterations are executed asynchronously on a dedicated CUDA stream followed by synchronization.

\textbf{Timing protocol.} Latency is measured using CUDA events recorded on the same CUDA stream. We record a start event, run inference for 1000 iterations back-to-back, record an end event, and synchronize on the end event. The average per-image latency (ms) is computed by
\begin{equation}
\mathrm{Latency} = \frac{t_{\mathrm{elapsed}}}{N},
\end{equation}
where $t_{\mathrm{elapsed}}$ is the CUDA-event elapsed time (ms) and $N$ is the number of runs. Throughput is reported as $\mathrm{FPS} = 1000/\mathrm{Latency}$.

\section{Discussion on Content-Driven Angle Estimation}
\label{app:why_decouple_angle}

\subsubsection{Why Angle-Coupled Positional Queries Can Be Ill-Posed}

We first discuss the \emph{representation-level} issue of coupling angle into positional queries.
Let the physical state space of an oriented bounding box be
\[
\mathcal{M}\triangleq\mathbb{R}^4 \times \big(S^1/(\theta\sim\theta+\pi)\big),
\]
where $(c_x,c_y,w,h)\in\mathbb{R}^4$, and $\theta$ is defined only up to a $\pi$ rotation for rectangles under the long-side definition; the same argument applies analogously to other common angle parameterizations.
Consider a DETR-style positional query
\[
Q_{\mathrm{pos}} = f(c_x,c_y,w,h,\theta), \qquad
f:\mathbb{R}^4\times[0,\pi)\to\mathbb{R}^d,
\]
where $\theta$ is represented by a scalar chart and $f$ is an Euclidean network.

\paragraph{(1) Quotient-consistency is required.}
Since $\theta$ and $\theta+\pi$ correspond to the same physical rectangle, a well-defined representation on $\mathcal{M}$ should satisfy
\[
f(x,\theta)=f(x,\theta+\pi),
\]
with $x=(c_x,c_y,w,h)$ and $\theta+\pi$ understood modulo $\pi$.
Equivalently, the seam endpoints of $[0,\pi)$ should map continuously:
\[
\lim_{\varepsilon\to 0^+} f(x,\varepsilon)
=
\lim_{\varepsilon\to 0^+} f(x,\pi-\varepsilon).
\]
Otherwise, two physically adjacent states near the seam may produce separated query vectors, making attention or matching overly sensitive to small angular perturbations.

\paragraph{(2) Standard scalar-angle constructions do not enforce this property.}
For common designs such as $\mathrm{MLP}\circ\mathrm{PE}$ with a scalar angle channel, the invariance above is generally \emph{not guaranteed} unless the encoding is explicitly made $\pi$-periodic.
Thus, angle-coupled positional queries can suffer from seam inconsistency and a mismatch between Euclidean regression and the periodic quotient geometry of OBBs.

\paragraph{(3) Long-side canonicalization does not fully remove instability.}
Although the long-side convention removes the $(w,h,\theta)\leftrightarrow(h,w,\theta+\frac{\pi}{2})$ ambiguity for clearly non-square instances, it becomes ill-conditioned near $w\approx h$.
A tiny perturbation may flip the long/short-side assignment and induce an approximately $\frac{\pi}{2}$ jump in the canonical angle.
If $Q_{\mathrm{pos}}$ depends on $\theta$, this instability is directly injected into the positional prior, which may hurt attention, matching, and optimization for near-square objects.
Moreover, model predictions are not guaranteed to satisfy the canonical constraint during training, so nearly equivalent parameterizations may still yield unstable gradients.

\paragraph{Implication.}
A simple robust remedy is to exclude $\theta$ from the positional query:
\[
Q_{\mathrm{pos}}=\phi(c_x,c_y,w,h).
\]
This prevents angular ambiguity from contaminating the geometric localization prior.
However, this only explains why \emph{naive} angle coupling is unstable.
One may still ask whether a better periodic or quotient-consistent encoding is sufficient.
The next subsection explains why the answer is still negative.

\subsubsection{Why Decoupling Angle from Positional Queries Is Preferred}

The above discussion concerns \emph{representation validity}, but the deeper issue is \emph{information decomposition}.
Even if angle is encoded in a periodic or quotient-consistent way, it is still not ideal to inject it into the positional branch.

The reason is that $(c_x,c_y,w,h)$ and $\theta$ play different roles.
The variables $(c_x,c_y,w,h)$ are geometric localization variables: they are spatially grounded and can reliably guide cross-attention before rich semantic evidence is aggregated.
By contrast, the correct orientation is often a \emph{content-disambiguated} variable, depending on appearance cues such as texture flow, object heading, part arrangement, and annotation convention.
Hence, angle is not merely another coordinate to be encoded in the positional prior.

Suppose the decoder query is
\begin{equation}
Q = Q_{\mathrm{content}} + Q_{\mathrm{pos}}(c_x,c_y,w,h,\hat{\theta}),
\end{equation}
where $\hat{\theta}$ is the current angle estimate.
If $\hat{\theta}$ is inaccurate in early training or decoding, then the model uses an orientation-conditioned positional prior \emph{before} sufficient content evidence has been collected.
As a result, feature aggregation is biased by an unreliable angle prior exactly when orientation is most ambiguous.

This also explains why alternative encodings do not fully solve the problem.
Encodings such as $\left[\sin\theta,\cos\theta\right]$, unit-vector parameterizations, or quotient-consistent embeddings such as $\left[\sin 2\theta,\cos 2\theta\right]$ may improve \emph{how} angle is represented, but they do not change \emph{where} angle is inferred.
They still route orientation-dependent information through the positional branch, rather than letting it emerge from content.

Therefore, we adopt the following principle: the positional query should encode only variables that are both spatially grounded and reliably available before content aggregation.
Under this principle, $(c_x,c_y,w,h)$ belong to $Q_{\mathrm{pos}}$, while $\theta$ should be inferred primarily from $Q_{\mathrm{content}}$.
Our decoupled formulation is thus not merely a workaround for scalar-angle ill-posedness, but a better information decomposition for oriented detection.

Our claims above are further proved by Sec. 4.6 in the main text, Appendix \ref{app:more_t_sne} and \ref{app:abl_pos}.

\section{Details on Decoupled Periodic Refinement}
\label{app:C_dpr}

\paragraph{Problem setting.}
We parameterize an OBB as $(c_x,c_y,w,h,\theta)$ with $\theta\in[0,\pi)$ due to the $\pi$-periodic symmetry of rectangles.
Standard DETR-style iterative refinement (e.g., inverse-sigmoid additive updates) is natural for Euclidean variables $(c_x,c_y,w,h)$, but applying the same \emph{unbounded Euclidean} regression/update to $\theta$ is ill-posed since angles live on a periodic domain with the seam $0\equiv\pi$.

\paragraph{Why Euclidean regression can contradict periodic geometry.}
Consider $\theta_{\mathrm{tgt}}=\varepsilon$ and $\theta_{\mathrm{pred}}=\pi-\varepsilon$ with $\varepsilon\!\rightarrow\!0^+$.
They are geometrically adjacent on $[0,\pi)$, yet the Euclidean gap is $\lvert \theta_{\mathrm{pred}}-\theta_{\mathrm{tgt}}\rvert\approx \pi$.
Thus, a Euclidean $L_1$ objective may encourage traversing the \emph{long arc} in the chart, causing seam instability and slowing early optimization (especially for near-square/ambiguous instances).

\paragraph{Periodic shortest-path operator.}
We define a $\pi$-periodic wrap and the signed shortest angular difference.
For any $u\in\mathbb{R}$,
\begin{equation}
\mathrm{wrap}_{\pi}(u)=u-\pi\left\lfloor \frac{u}{\pi}\right\rfloor \in [0,\pi).
\label{eq:dpr_wrap}
\end{equation}
Then
\begin{equation}
\Delta_{\pi}(\theta_p,\theta_t)
=
\mathrm{wrap}_{\pi}\!\left(\theta_p-\theta_t+\tfrac{\pi}{2}\right)-\tfrac{\pi}{2}
\in \left[-\tfrac{\pi}{2},\tfrac{\pi}{2}\right).
\label{eq:dpr_shortest_delta}
\end{equation}
This is seam-consistent: crossing $0/\pi$ does not create a large jump in $\Delta_{\pi}$ (a.e.).

\paragraph{Shortest-path periodic $L_1$ for angle regression.}
\begin{equation}
\mathcal{L}_{\angle}
=
\left|\Delta_{\pi}(\theta_{\mathrm{pred}},\theta_{\mathrm{tgt}})\right|.
\label{eq:dpr_sp_l1}
\end{equation}
Compared with Euclidean $L_1$, Eq.~\eqref{eq:dpr_sp_l1} measures the geodesic error on the $\pi$-periodic domain and yields shortest-path gradients.

\paragraph{Bounded periodic refinement update.}
We keep inverse-sigmoid refinement for $(c_x,c_y,w,h)$, and decouple $\theta$ with a bounded periodic update:
\begin{equation}
\theta^{(i+1)}
=
\mathrm{wrap}_{\pi}\!\Big(\theta^{(i)} + \alpha_i \cdot g(\delta^{(i)})\Big),
\label{eq:dpr_update}
\end{equation}
where $\delta^{(i)}$ is the raw angle offset at decoder layer $i$, $g(\cdot)$ is bounded (we use $\tanh$), and
\begin{equation}
\alpha_i = \alpha_0^{-i}, \qquad i\in\{0,1,\dots,L-1\}.
\label{eq:dpr_alpha}
\end{equation}
This yields bounded steps and a coarse-to-fine refinement across layers.

\section{Detailed Results on Benchmarks}

\begin{table*}[t]
\centering
\caption{Comparison on DOTA-1.0 under single-scale training and testing protocol. Left: high-precision comparison. Right: results of RiO-DETR on other backbones.}
\label{tab:combined_tables}

\begin{minipage}[t]{0.455\linewidth}
\centering
\setlength{\tabcolsep}{5pt}
\renewcommand{\arraystretch}{1.1}
\resizebox{\linewidth}{!}{%
\begin{tabular}{lccc}
\toprule
\textbf{Method} & \textbf{Backbone} & \textbf{AP$_{50}$} & \textbf{AP$_{75}$} \\
\midrule
Oriented R-CNN~\cite{xie2021oriented} & R-50 & 74.19 & 46.96 \\
RoI Trans.~\cite{ding2019learning} & R-50 & 74.05 & 46.54 \\
RoI Trans.~\cite{ding2019learning} & Swin-T & 76.49 & 50.15 \\
ReDet~\cite{han2021redet} & ReR-50 & 76.25 & 50.86 \\
ARS-DETR~\cite{zeng2024ars} & R-50 & 73.79 & 49.01 \\
ARS-DETR~\cite{zeng2024ars} & Swin-T & 75.79 & 51.11 \\
RHINO w/ KLD \cite{lee2025hausdorff} & R-50 & 78.49 & 51.84 \\
RHINO w/ GRIoU \cite{lee2025hausdorff} & R-50 & 77.24 & 53.91 \\
YOLO26x-obb & YOLO26x & 80.12 & 58.04 \\
\midrule
\textbf{RiO-DETR-x w/ KLD} & HGNet-B5 & \textbf{81.78} & \underline{56.35} \\
\textbf{RiO-DETR-x w/ GRIoU} & HGNet-B5 & \underline{80.62} & \textbf{58.49} \\
\bottomrule
\end{tabular}
}
\label{tab:hp}
\end{minipage}
\hfill
\begin{minipage}[t]{0.48\linewidth}
\centering
\setlength{\tabcolsep}{4pt}
\renewcommand{\arraystretch}{1.1}
\resizebox{\linewidth}{!}{
\begin{tabular}{llcccc}
\toprule
\textbf{Methods} & \textbf{Backbone} & \textbf{\#P} & \textbf{Flops} & \textbf{Lat.} & \textbf{AP$_{50}$} \\
\midrule
LSKNet-S \cite{li2025lsknet} & LSKNet-S & 31.0M & 161G & 146.2 & 77.5 \\
AO2-DETR \cite{dai2022ao2} & R-50 & 74.3M & 304G & -- & 77.7 \\
RHINO-DETR \cite{lee2025hausdorff} & R-50 & 47.6M & 566G & 47.8 & 78.7 \\
RHINO-DETR \cite{lee2025hausdorff} & Swin-T & 50.8M & 609G & 57.0 & 79.4 \\
Oriented-DETR \cite{zhao2024projecting} & R-50 & 57.2M & 302G & 87.2 & 79.1 \\
Oriented-DETR \cite{zhao2024projecting} & Swin-T & 57.7M & 309G & 89.8 & 79.8 \\
YOLO26m-obb \cite{sapkota2025yolo26} & YOLO26m & 21.2M & 183G & 10.2 & 80.0 \\
YOLO26l-obb \cite{sapkota2025yolo26} & YOLO26l & 25.6M & 230G & 13.0 & 80.2 \\
YOLO26x-obb \cite{sapkota2025yolo26} & YOLO26x & 57.6M & 517G & 30.5 & 80.4 \\
\midrule
\textbf{RiO-DETR-R50} & R-50 & 52.7M & 217G & 12.6 & 80.1 \\
\textbf{RiO-DETR-Swin-T} & Swin-T & 55.9M & 221G & 28.5 & 80.8 \\
\textbf{RiO-DETR-LSKNet} & LSKNet-S & 42.1M & 176G & 88.7 & 81.2 \\
\bottomrule
\end{tabular}
}
\label{table:dota_ss_compare}
\end{minipage}
\end{table*}

\begin{table}[t]
\centering

\begin{minipage}{\linewidth}
\centering
\caption{Comparison with state-of-the-art oriented object detectors on DOTA-1.0 under multi-scale training and testing protocol.}
\label{table:dota_ms_compare}

\setlength{\tabcolsep}{3pt}
\resizebox{\linewidth}{!}{
\begin{tabular}{ll*{19}{c}}
\toprule
\textbf{Methods} & \textbf{Backbone} & \textbf{\#P} & \textbf{Flops} & \textbf{Lat.} & \textbf{AP$_{50}$} &
PL & BD & BR & GTF & SV & LV & SH & TC & BC & ST & SBF & RA & HA & SP & HC \\
\midrule

\multicolumn{21}{c}{\textbf{Non-Real-time Oriented Object Detectors}} \\
\midrule
PKINet-S \cite{cai2024poly} & PKINet-S & 30.8M & 190G & 187.5 & 81.06 & 89.0 & 86.7 & 59.0 & 81.2 & 80.4 & 84.9 & 88.1 & 90.9 & 86.6 & 87.3 & 67.1 & 74.8 & 78.2 & 81.9 & 70.6 \\
LSKNet-S \cite{li2025lsknet} & LSKNet-S & 31.0M & 161G & 146.2 & 81.64 & 89.6 & 86.3 & 63.1 & 83.7 & 82.2 & 86.1 & 88.7 & 90.9 & 88.4 & 87.4 & 71.7 & 69.6 & 78.9 & 81.8 & 76.5 \\
Strip R-CNN-S \cite{yuan2025strip} & StripNet-S & 30.5M & 159G & 106.3 & 82.28 & 89.2 & 85.6 & 62.4 & 83.7 & 81.9 & 86.6 & 88.8 & 90.9 & 88.0 & 87.9 & 72.1 & 71.9 & 79.2 & 82.5 & 82.8 \\
\midrule

\multicolumn{21}{c}{\textbf{Real-time Oriented Object Detectors}} \\
\midrule
PPYOLOE-R-x \cite{xu2022pp} & CRN-x & 98.4M & 264G & 32.8 & 80.73 & 88.5 & 84.5 & 60.6 & 77.7 & 83.3 & 85.4 & 89.0 & 90.8 & 88.5 & 87.5 & 69.3 & 66.0 & 77.9 & 81.4 & 80.9 \\
RTMDet-R-l \cite{lyu2022rtmdet} & CSPNext-l & 52.3M & 205G & 23.7 & 81.33 & 88.4 & 85.0 & 57.3 & 80.5 & 80.6 & 84.9 & 88.1 & 90.9 & 86.3 & 87.6 & 69.3 & 70.6 & 78.6 & 81.0 & 79.2 \\
YOLO26m-obb \cite{sapkota2025yolo26} & YOLO26l & 25.6M & 230G & 10.2 & 81.00 & 89.1 & 86.1 & 59.1 & 78.1 & 81.2 & 85.7 & 88.7 & 90.9 & 87.7 & 88.9 & 70.6 & 69.3 & 78.8 & 82.4 & 83.3 \\
YOLO26x-obb \cite{sapkota2025yolo26} & YOLO26x & 57.6M & 517G & 30.5 & 81.70 & 89.2 & 86.9 & 62.0 & 80.7 & 81.6 & 86.0 & 88.7 & 90.8 & 87.2 & 88.8 & 70.1 & 68.1 & 84.5 & 81.7 & 88.1 \\
\textbf{RiO-DETR-m} & HGNetv2-B2 & 18.6M & 158G & 8.8 & 81.49 & 87.3 & 87.2 & 61.7 & 81.8 & 82.5 & 86.2 & 89.2 & 90.6 & 88.7 & 88.7 & 74.5 & 71.4 & 78.8 & 74.4 & 79.7 \\
\textbf{RiO-DETR-x} & HGNetv2-B5 & 62.5M & 527G & 29.9 & 81.76 & 83.8 & 87.3 & 62.6 & 79.9 & 82.7 & 86.3 & 89.1 & 90.6 & 88.5 & 88.0 & 72.5 & 75.2 & 78.9 & 81.8 & 79.2 \\
\bottomrule
\end{tabular}
}
\end{minipage}
\end{table}

\begin{table*}[t]
\centering
\caption{Comparison with state-of-the-art models on the FAIR-1M-2.0 dataset.
The object categories in the table of C1--C37 (in order) is:
Boeing737, Boeing747, Boeing777, Boeing787, C919, A220, A321, A330, A350, ARJ21, OtherAirplane, SmallCar, Bus, CargoTruck, DumpTruck, Van, Trailer, Tractor, Excavator, TruckTractor, OtherVehicle, BasketballCourt, TennisCourt, FootballField, BaseballField, Intersection, Roundabout, Bridge, PassengerShip, Motorboat, FishingBoat, Tugboat, EngineeringShip, LiquidCargoShip, DryCargoShip, Warship, OtherShip.}
\label{tab:fair1m_ms_compare}
\scriptsize
\setlength{\tabcolsep}{2.8pt}
\renewcommand{\arraystretch}{1.08}
\resizebox{\linewidth}{!}{
\begin{tabular}{lcccccccccccccccccccc}
\toprule
\textbf{Method}
& \makecell{\textbf{C1}\\\textbf{C20}}
& \makecell{\textbf{C2}\\\textbf{C21}}
& \makecell{\textbf{C3}\\\textbf{C22}}
& \makecell{\textbf{C4}\\\textbf{C23}}
& \makecell{\textbf{C5}\\\textbf{C24}}
& \makecell{\textbf{C6}\\\textbf{C25}}
& \makecell{\textbf{C7}\\\textbf{C26}}
& \makecell{\textbf{C8}\\\textbf{C27}}
& \makecell{\textbf{C9}\\\textbf{C28}}
& \makecell{\textbf{C10}\\\textbf{C29}}
& \makecell{\textbf{C11}\\\textbf{C30}}
& \makecell{\textbf{C12}\\\textbf{C31}}
& \makecell{\textbf{C13}\\\textbf{C32}}
& \makecell{\textbf{C14}\\\textbf{C33}}
& \makecell{\textbf{C15}\\\textbf{C34}}
& \makecell{\textbf{C16}\\\textbf{C35}}
& \makecell{\textbf{C17}\\\textbf{C36}}
& \makecell{\textbf{C18}\\\textbf{C37}}
& \makecell{\textbf{C19}\\\textbf{--}}
& \makecell{\textbf{AP$_{50}$}\\\textbf{(\%)}} \\
\midrule

SASM RepPoints$^{*}$ & \makecell{38.3\\0.1} & \makecell{61.7\\2.4} & \makecell{15.0\\37.2} & \makecell{35.4\\80.4} & \makecell{0.4\\49.9} & \makecell{44.0\\87.4} & \makecell{53.3\\44.5} & \makecell{30.6\\54.8} & \makecell{20.7\\28.3} & \makecell{3.9\\14.7} & \makecell{70.7\\31.9} & \makecell{53.6\\14.3} & \makecell{9.4\\24.8} & \makecell{29.9\\17.7} & \makecell{21.2\\25.7} & \makecell{48.3\\48.4} & \makecell{2.8\\32.2} & \makecell{0.1\\5.9} & \makecell{2.2\\--} & 30.9 \\
\midrule
R-FCOS$^{*}$ & \makecell{39.2\\5.4} & \makecell{82.7\\2.7} & \makecell{15.9\\43.7} & \makecell{52.0\\86.2} & \makecell{0.2\\56.3} & \makecell{41.0\\88.1} & \makecell{54.5\\48.0} & \makecell{44.2\\57.2} & \makecell{43.5\\25.7} & \makecell{9.3\\24.0} & \makecell{73.1\\43.8} & \makecell{55.1\\22.0} & \makecell{11.5\\18.9} & \makecell{37.9\\31.5} & \makecell{26.0\\33.4} & \makecell{51.0\\56.8} & \makecell{6.3\\27.3} & \makecell{1.1\\7.3} & \makecell{12.9\\--} & 36.1 \\
\midrule
S2A-Net$^{*}$ & \makecell{40.4\\2.5} & \makecell{83.7\\2.2} & \makecell{17.2\\46.2} & \makecell{49.9\\82.1} & \makecell{0.2\\59.0} & \makecell{42.5\\88.2} & \makecell{57.0\\41.5} & \makecell{42.7\\62.9} & \makecell{52.0\\20.3} & \makecell{10.3\\24.8} & \makecell{74.8\\42.9} & \makecell{66.2\\19.9} & \makecell{13.4\\24.8} & \makecell{39.7\\32.5} & \makecell{35.7\\33.3} & \makecell{61.7\\58.5} & \makecell{2.8\\34.9} & \makecell{0.8\\6.6} & \makecell{10.5\\--} & 37.4 \\
\midrule
R-Faster R-CNN$^{*}$ & \makecell{37.5\\11.9} & \makecell{82.5\\2.0} & \makecell{15.4\\42.6} & \makecell{47.7\\83.3} & \makecell{9.4\\55.4} & \makecell{41.7\\89.4} & \makecell{50.8\\50.4} & \makecell{47.5\\65.1} & \makecell{60.6\\26.1} & \makecell{8.3\\19.0} & \makecell{72.5\\47.5} & \makecell{57.7\\17.9} & \makecell{12.9\\24.2} & \makecell{41.4\\26.3} & \makecell{38.2\\30.5} & \makecell{53.4\\54.1} & \makecell{11.7\\26.3} & \makecell{2.1\\8.0} & \makecell{16.8\\--} & 37.5 \\
\midrule
O-RepPoints$^{*}$ & \makecell{37.7\\5.7} & \makecell{81.4\\0.9} & \makecell{15.4\\46.3} & \makecell{48.8\\84.3} & \makecell{7.9\\61.4} & \makecell{43.4\\89.2} & \makecell{54.6\\46.7} & \makecell{43.5\\60.8} & \makecell{56.6\\33.2} & \makecell{9.5\\24.6} & \makecell{73.3\\50.4} & \makecell{62.9\\24.9} & \makecell{15.4\\22.2} & \makecell{42.0\\32.8} & \makecell{42.8\\34.7} & \makecell{59.5\\60.0} & \makecell{4.5\\37.9} & \makecell{0.6\\7.4} & \makecell{17.0\\--} & 38.9 \\
\midrule
O-RCNN$^{*}$ & \makecell{37.6\\10.4} & \makecell{84.8\\1.2} & \makecell{16.9\\47.2} & \makecell{49.0\\83.3} & \makecell{7.3\\59.6} & \makecell{41.6\\89.1} & \makecell{49.7\\48.3} & \makecell{43.1\\62.6} & \makecell{62.4\\31.8} & \makecell{13.1\\26.6} & \makecell{72.1\\58.5} & \makecell{59.8\\28.2} & \makecell{18.5\\25.5} & \makecell{43.9\\38.8} & \makecell{41.5\\37.3} & \makecell{56.0\\64.6} & \makecell{7.9\\38.7} & \makecell{2.2\\10.7} & \makecell{24.4\\--} & 40.4 \\
\midrule
RoI Trans.$^{*}$ & \makecell{38.9\\16.5} & \makecell{81.8\\1.3} & \makecell{14.6\\50.4} & \makecell{46.1\\86.2} & \makecell{7.1\\61.0} & \makecell{42.4\\88.7} & \makecell{55.4\\50.2} & \makecell{38.6\\66.3} & \makecell{57.3\\30.7} & \makecell{10.3\\26.3} & \makecell{73.4\\52.1} & \makecell{62.2\\27.7} & \makecell{20.6\\26.9} & \makecell{43.7\\32.2} & \makecell{43.4\\32.6} & \makecell{58.5\\61.9} & \makecell{13.7\\37.0} & \makecell{1.9\\9.3} & \makecell{21.0\\--} & 40.2 \\
\midrule
LOOD (RT)$^{*}$ & \makecell{44.6\\22.2} & \makecell{83.4\\1.7} & \makecell{18.5\\52.3} & \makecell{47.9\\84.7} & \makecell{11.7\\67.3} & \makecell{44.6\\89.8} & \makecell{59.4\\50.8} & \makecell{47.5\\68.2} & \makecell{59.9\\37.3} & \makecell{18.0\\28.8} & \makecell{76.1\\51.9} & \makecell{63.3\\28.3} & \makecell{20.2\\24.6} & \makecell{45.0\\33.9} & \makecell{45.5\\39.0} & \makecell{59.1\\64.1} & \makecell{10.4\\41.4} & \makecell{2.2\\9.7} & \makecell{21.7\\--} & 42.6 \\
\midrule
ReDet$^{*}$ & \makecell{37.1\\16.7} & \makecell{86.6\\1.4} & \makecell{16.5\\55.8} & \makecell{57.2\\86.2} & \makecell{9.4\\66.8} & \makecell{44.9\\91.0} & \makecell{59.1\\51.1} & \makecell{45.4\\74.5} & \makecell{60.3\\37.6} & \makecell{11.1\\31.8} & \makecell{76.4\\57.2} & \makecell{63.7\\33.8} & \makecell{20.9\\21.3} & \makecell{44.2\\45.5} & \makecell{42.7\\36.5} & \makecell{59.7\\67.0} & \makecell{13.2\\39.9} & \makecell{1.8\\10.8} & \makecell{24.2\\--} & 43.2 \\
\midrule
LOOD (RD)$^{*}$ & \makecell{43.1\\14.9} & \makecell{90.6\\1.6} & \makecell{22.2\\54.2} & \makecell{55.8\\84.6} & \makecell{9.4\\64.7} & \makecell{46.6\\92.4} & \makecell{63.7\\50.7} & \makecell{55.8\\71.7} & \makecell{68.3\\41.2} & \makecell{16.5\\34.1} & \makecell{77.0\\56.5} & \makecell{64.9\\34.6} & \makecell{23.3\\22.6} & \makecell{45.8\\46.3} & \makecell{45.5\\38.8} & \makecell{61.2\\68.6} & \makecell{12.5\\48.0} & \makecell{4.8\\9.6} & \makecell{19.0\\--} & 44.9 \\
\midrule
Strip-RCNN-S & \makecell{35.2\\54.7} & \makecell{80.1\\5.6} & \makecell{11.9\\51.1} & \makecell{45.3\\79.5} & \makecell{19.8\\59.9} & \makecell{38.9\\87.2} & \makecell{52.4\\48.5} & \makecell{43.5\\55.4} & \makecell{64.2\\39.7} & \makecell{10.7\\27.1} & \makecell{68.3\\61.4} & \makecell{73.7\\45.4} & \makecell{31.8\\22.0} & \makecell{53.9\\42.3} & \makecell{54.4\\37.7} & \makecell{72.9\\67.0} & \makecell{26.3\\49.0} & \makecell{4.6\\17.1} & \makecell{37.4\\--} & 45.3 \\
\midrule
LSKNet-S & \makecell{31.9\\56.6} & \makecell{80.0\\10.8} & \makecell{4.7\\49.9} & \makecell{38.3\\79.5} & \makecell{19.7\\62.6} & \makecell{42.3\\86.9} & \makecell{43.1\\50.8} & \makecell{39.1\\65.4} & \makecell{70.1\\46.3} & \makecell{14.3\\28.7} & \makecell{68.5\\58.3} & \makecell{74.2\\42.6} & \makecell{40.6\\26.0} & \makecell{55.4\\40.9} & \makecell{54.7\\38.1} & \makecell{73.4\\66.5} & \makecell{24.5\\52.7} & \makecell{1.0\\15.2} & \makecell{40.4\\--} & 45.8 \\
\midrule
YOLO26m-obb & \makecell{28.2\\47.8} & \makecell{80.7\\6.9} & \makecell{3.7\\46.8} & \makecell{38.6\\78.1} & \makecell{16.5\\57.9} & \makecell{40.8\\85.7} & \makecell{41.7\\46.8} & \makecell{40.7\\62.0} & \makecell{60.7\\35.7} & \makecell{8.7\\29.0} & \makecell{66.7\\55.0} & \makecell{71.6\\41.0} & \makecell{31.9\\24.2} & \makecell{52.3\\38.1} & \makecell{50.7\\34.7} & \makecell{71.3\\64.7} & \makecell{18.9\\50.0} & \makecell{0.1\\15.1} & \makecell{29.2\\--} & 42.5 \\
\midrule
YOLO26x-obb & \makecell{34.2\\54.6} & \makecell{83.3\\8.2} & \makecell{7.5\\50.5} & \makecell{44.3\\80.5} & \makecell{23.6\\61.9} & \makecell{43.4\\87.8} & \makecell{45.7\\51.2} & \makecell{45.2\\67.1} & \makecell{65.8\\41.2} & \makecell{14.5\\33.4} & \makecell{69.7\\60.0} & \makecell{74.4\\47.1} & \makecell{37.1\\26.0} & \makecell{56.4\\43.6} & \makecell{54.8\\39.6} & \makecell{74.2\\68.0} & \makecell{25.3\\53.9} & \makecell{1.0\\17.7} & \makecell{35.2\\--} & 46.7 \\
\midrule
\textbf{RiO-DETR-m} & \makecell{36.0\\40.4} & \makecell{82.4\\6.0} & \makecell{9.6\\45.1} & \makecell{44.9\\78.9} & \makecell{20.5\\56.9} & \makecell{37.4\\85.6} & \makecell{46.1\\51.6} & \makecell{53.0\\67.2} & \makecell{54.6\\30.3} & \makecell{13.5\\34.7} & \makecell{68.7\\62.6} & \makecell{72.7\\44.4} & \makecell{22.4\\25.3} & \makecell{51.9\\41.3} & \makecell{49.5\\38.8} & \makecell{73.1\\65.8} & \makecell{17.7\\42.6} & \makecell{0.7\\12.4} & \makecell{27.2\\--} & \textbf{43.6} \\
\midrule
\textbf{RiO-DETR-x} & \makecell{35.4\\53.7} & \makecell{85.2\\6.8} & \makecell{9.0\\50.6} & \makecell{48.2\\80.9} & \makecell{26.0\\61.0} & \makecell{43.5\\88.2} & \makecell{47.0\\54.0} & \makecell{49.3\\71.0} & \makecell{61.7\\39.7} & \makecell{14.0\\36.6} & \makecell{69.9\\60.8} & \makecell{75.2\\50.0} & \makecell{35.1\\25.4} & \makecell{57.8\\45.2} & \makecell{55.5\\40.2} & \makecell{75.5\\68.7} & \makecell{26.7\\54.3} & \makecell{1.6\\19.2} & \makecell{31.8\\--} & \textbf{47.4} \\
\bottomrule
\end{tabular}
}

\vspace{-0.2cm}
\end{table*}

\subsection{Per-class Results on DOTA-1.0 Multi-scale}

We provide the per-class results on DOTA-1.0 under the multi-scale training/testing protocol in Table~\ref{table:dota_ms_compare}. The table reports detailed AP$_{50}$ scores for each category. These results supplement the overall comparisons in the main text with a category-level evaluation under the multi-scale setting.

\subsection{Per-class Results on FAIR-1M-2.0 Multi-scale}

We provide the per-class results on FAIR-1M-2.0 under the multi-scale setting in Table~\ref{tab:fair1m_ms_compare}. The table presents detection results for all categories and supplements the overall quantitative comparisons in the main text by providing a class-wise evaluation of RiO-DETR-m, RiO-DETR-x, and other compared methods.

\subsection{Results of High-Precision Metrics}

In mainstream oriented object detection benchmarks, AP$_{50}$ is commonly used as the only metric for performance comparison. We follow this convention and use AP$_{50}$ as the main evaluation criterion. To further assess localization accuracy under stricter matching conditions, we additionally report AP$_{75}$.

Our default model adopts KLD as part of the regression loss, since it is a widely used and stable choice in rotated object detection, providing a fair and standard setting for comparison with prior methods. Under this default setting, RiO-DETR-x achieves the best AP$_{50}$ of \textbf{81.78}, surpassing all previous methods.

Lee et al. \cite{lee2025hausdorff} found that replacing KLD with GRIoU can shift the optimization more toward overlap-aware high-precision localization in oriented DETRs. We follow their setting, and RiO-DETR-x w/ GRIoU achieves \textbf{80.62} AP$_{50}$ and \textbf{58.49} AP$_{75}$, outperforming all compared methods on both metrics.

\subsection{Results on Other Backbones}
\label{app:other_backbones}

Table~\ref{table:dota_ss_compare} reports the single-scale training and testing results of RiO-DETR when instantiated with different backbones on DOTA-1.0. The base model size is x.
We keep the detector head and training protocol unchanged and only replace the backbone, so that the comparison reflects the portability of the proposed oriented DETR design.
Across all tested backbones, RiO-DETR consistently achieves better accuracy than compared DETR-based and CNN-based models, while maintaining a favorable efficiency profile.


\begin{figure}[t]
\centering
\begin{minipage}{0.53\linewidth}
  \centering
  \includegraphics[width=\linewidth]{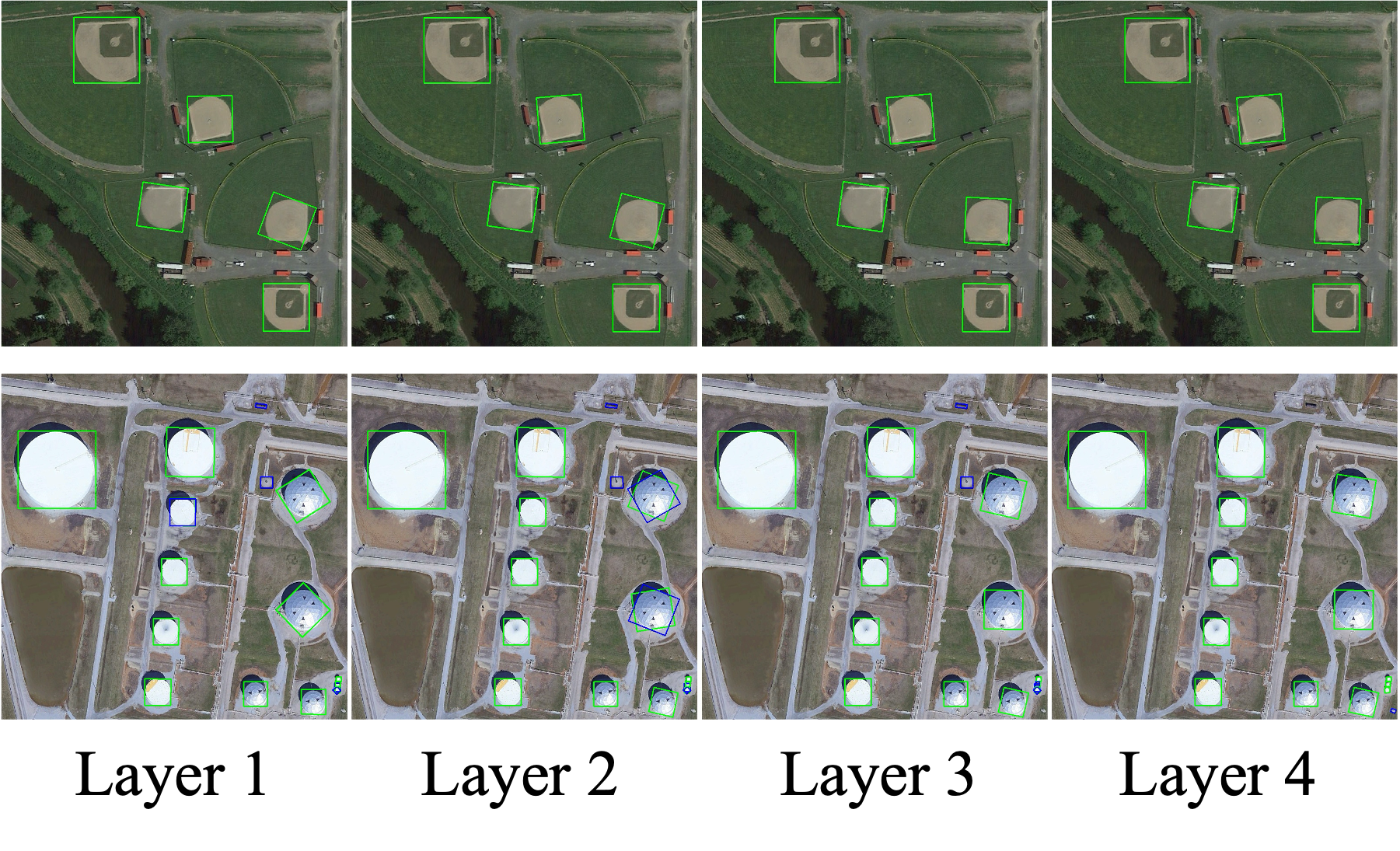}
  \caption{Visualizations of layer-wise angular refinement for square-like instances.}
  \label{fig:angle_more}
\end{minipage}
\hfill
\begin{minipage}{0.45\linewidth}
  \centering
  \includegraphics[width=\linewidth]{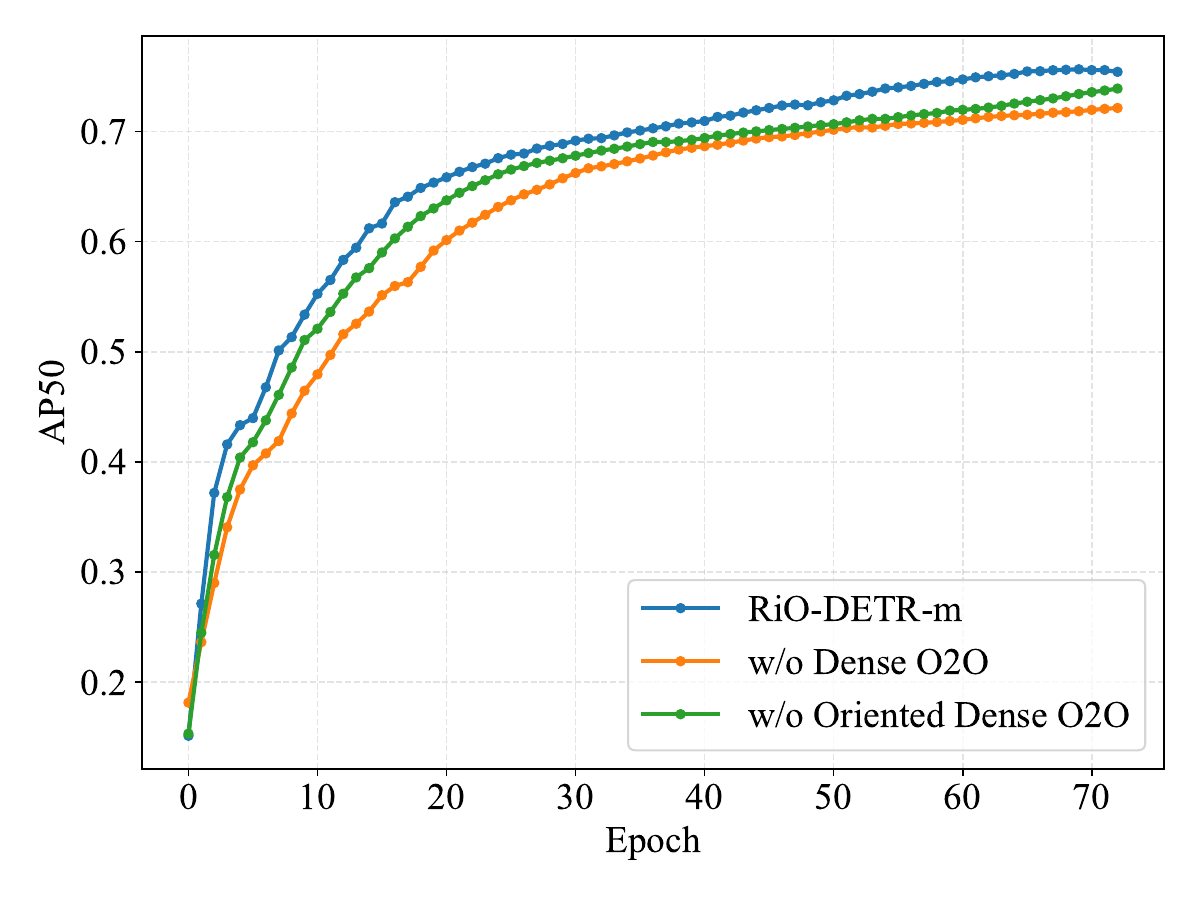}
  \caption{Per-epoch AP$_{50}$, which shows the difference in convergence speed.}
  \label{fig:o2o_ap50}
\end{minipage}
\end{figure}

\section{Further Analyses and Visualizations}

\subsection{Further Visualization of Oriented Dense O2O}

Fig.~\ref{fig:o2o_ap50} shows the per-epoch AP$_{50}$ during training for three settings: without Dense O2O, with standard Dense O2O, and with our Oriented Dense O2O. 
Introducing Dense O2O already improves the convergence speed compared with the vanilla training scheme, indicating that denser one-to-one supervision provides stronger learning signals at early stages. 
Our Oriented Dense O2O further accelerates convergence and achieves consistently higher AP$_{50}$ throughout training. 
In particular, the gap becomes evident in the early epochs, suggesting that orientation-aware densification provides more effective angular supervision and allows the model to enter a stable refinement regime earlier. 

\subsection{Angular Prediction across Decoder Layers}

To further analyze the coarse-to-fine refinement behavior, we present a layer-wise evaluation of angular predictions in Table \ref{tab:layerwise_angle}. “Avg Error” denotes the mean absolute angular error w.r.t. ground truth, while “Avg Delta” measures the angular adjustment introduced by each decoder layer. As expected, Layer 1 produces the largest shift ($0.515^\circ$), establishing the primary orientation. In later layers, the “Avg Delta” decreases exponentially ($0.221^\circ \rightarrow 0.048^\circ$), indicating a transition to fine-grained refinements. This confirms that the decaying factor $\alpha_i$ stabilizes decoding, suppressing late-stage oscillations and enabling precise convergence. We also provide visualizations for square-like cases in Figure~\ref{fig:angle_more} to prove the robustness of our angle prediction.

\subsection{More t-SNE Visualizations}
\label{app:more_t_sne}

\begin{figure}[t]
  \centering
  \includegraphics[width=\linewidth, keepaspectratio]{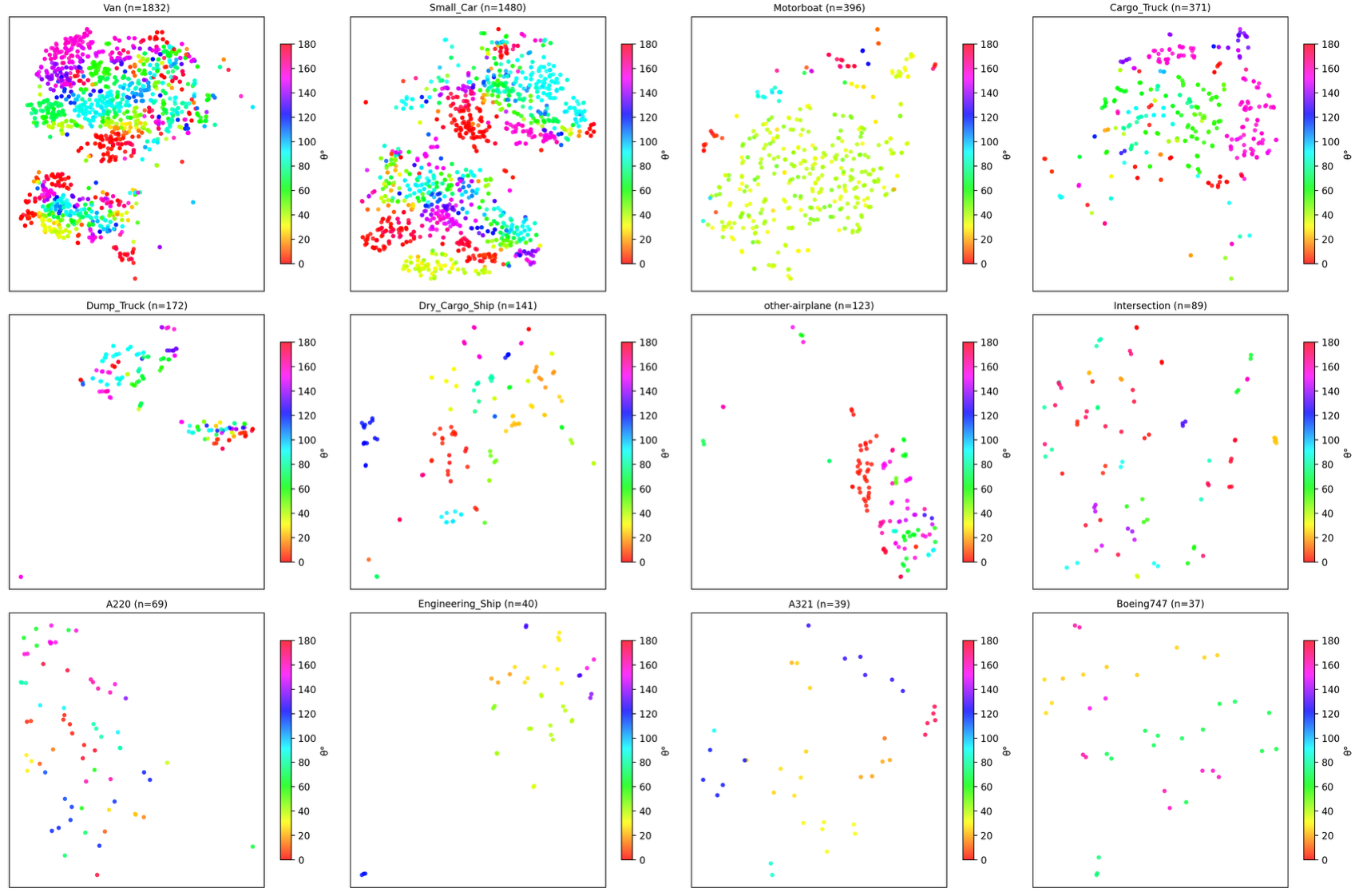}
  \caption{Per-category t-SNE visualizations with orientation coloring on FAIR-1M-2.0.}
  \label{fig:attention_vis}
\end{figure}

We provide additional t-SNE visualizations on FAIR-1M-2.0 to better understand the representation learned by the content queries. Figure~\ref{fig:attention_vis} presents per-category t-SNE plots, with points colored by the oriented bounding-box angle $\theta$.
Across many categories, the color distribution is not strictly separated into isolated angle-specific islands; instead, different angles are often interleaved within the same local neighborhoods. This indicates that the content-query space is not trivially dominated by the periodic angle signal, but retains content/appearance cues as the primary organizing factor.
Meanwhile, for categories with strong viewpoint regularities (e.g., vehicles on roads or ships aligned with shipping lanes), we occasionally observe mild angle-correlated gradients or sub-structures, which is expected given dataset bias and scene geometry.
Overall, these visualizations qualitatively support the design goal of learning orientation from content while avoiding degenerate representations that collapse to a single direction cue.

\begin{table}[!htbp] 
\centering
\tiny 

\begin{minipage}[t]{0.48\linewidth}
    \centering
    \caption{Hyperparameter ablations for Decoupled Periodic Refinement.}
    \label{tab:dpr_hyperparams}
    \begin{tabularx}{\linewidth}{p{5em} *{4}{>{\centering\arraybackslash}X}}
    \toprule
    \boldmath$\alpha_0$ & 1.0 & \textbf{1.5} & 2.0 & 2.5 \\
    AP$_{50}$ & 74.27 & \textbf{74.74} & 74.45 & 74.68 \\
    \midrule
    \end{tabularx}
    \begin{tabularx}{\linewidth}{p{5em} *{4}{>{\centering\arraybackslash}X}}
    \textbf{Strategy} & None & Linear & \textbf{Exp} & Power \\
    AP$_{50}$ & 74.18 & 74.49 & \textbf{74.74} & 74.61 \\
    \midrule
    \end{tabularx}
    \begin{tabularx}{\linewidth}{p{5em} *{4}{>{\centering\arraybackslash}X}}
    \textbf{Activation} & Linear & \textbf{Tanh} & Sin & Sigmoid\\
    AP$_{50}$ & 73.92 & \textbf{74.74} & 74.69 & 74.23 \\
    \bottomrule
    \end{tabularx}
\end{minipage}
\hfill 
\begin{minipage}[t]{0.48\linewidth}
    \centering
    \caption{Layer-wise analysis of angular prediction during decoding of RiO-x.}
    \label{tab:layerwise_angle}
    \begin{tabularx}{\linewidth}{*{3}{>{\centering\arraybackslash}X}}
    \toprule
    \textbf{Layer} & \textbf{Avg Error} ($^\circ$) & \textbf{Avg Delta} ($^\circ$) \\
    \midrule
    0 (Initial) & 8.20 & 0.000 \\
    1           & 8.11 & 0.515 \\
    2           & 8.09 & 0.221 \\
    3           & 8.08 & 0.048 \\
    \bottomrule
    \end{tabularx}
\end{minipage}

\vspace{0.6cm}

\begin{minipage}[t]{\linewidth}
\centering
\caption{Ablation on injecting angle into the positional query.}
\label{tab:angle_positional_ablation}
\begin{tabular}{lccccccc}
\toprule
\textbf{Positional Query Representation }
& \textbf{Angle in Positional Branch }
& \textbf{Periodicity-aware Encoding }
& \textbf{AP$_{50}$} \\
\midrule
$(c_x, c_y, w, h, \theta)$ 
& Yes 
& No 
& 73.47 \\
$(c_x, c_y, w, h, \sin\theta, \cos\theta)$ 
& Yes 
& Yes 
& 72.60 \\
$(c_x, c_y, w, h, \sin 2\theta, \cos 2\theta)$ 
& Yes 
& Yes 
& 73.52 \\
$(c_x, c_y, w, h)$ \textbf{(Ours)}
& No 
& N/A 
& 74.18 \\
\bottomrule
\end{tabular}
\end{minipage}

\end{table}

\section{Further Ablation Studies}

\subsubsection{Ablation on Positional Encodings.}
\label{app:abl_pos}

Table~\ref{tab:angle_positional_ablation} examines whether the gain of our design mainly comes from using a better periodic angle parameterization, or from decoupling angle from the positional branch altogether. Replacing the raw angle $\theta$ with $(\sin\theta,\cos\theta)$ even degrades performance from 73.47 to 72.60 AP$_{50}$, while the quotient-consistent encoding $(\sin 2\theta,\cos 2\theta)$ only brings a marginal improvement to 73.52 AP$_{50}$. In contrast, our fully decoupled design, which removes angle from the positional query and keeps only $(c_x,c_y,w,h)$, achieves the best result of 74.18 AP$_{50}$. These results suggest that the limitation is not merely caused by a suboptimal periodic encoding. Instead, injecting angle into the positional branch is itself less suitable for oriented DETR. Overall, this ablation supports our claim that angle should be primarily estimated from content features, rather than being encoded in the positional query.

\subsubsection{Hyperparameter Ablation on Decoupled Periodic Refinement.}

Table~\ref{tab:dpr_hyperparams} examines key hyperparameters in Decoupled Periodic Refinement. The decay base factor $\alpha_0=1.5$ yields the best result (74.74 AP$_{50}$), while smaller (1.0) or larger values (2.0--2.5) degrade performance, highlighting the need for a proper decay rate. Exponential decay consistently outperforms none, linear, and power schedules, and \texttt{tanh} achieves the highest AP$_{50}$ (74.74) among the tested activation functions.

\clearpage
\bibliographystyle{splncs04}
\bibliography{main}